\pdfoutput=1

\documentclass[11pt]{article}

\usepackage[final]{acl}

\usepackage{times}
\usepackage{latexsym}

\usepackage[T1]{fontenc}

\usepackage[utf8]{inputenc}

\usepackage{microtype}

\usepackage{inconsolata}

\usepackage{graphicx}
\usepackage{float}
\usepackage{amsfonts}
\usepackage{amsmath}
\usepackage{hyperref}

\makeatletter
\newcommand{\blfootnote}[1]{%
  \begingroup
  \renewcommand\thefootnote{}\footnote{#1}%
  \addtocounter{footnote}{-1}%
  \endgroup
}
\makeatother
\title{Addition in Four Movements: Mapping Layer-wise Information Trajectories in LLMs}

\author{
Yao Yan\\
Chongqing Normal University\\
Chongqing, China\\
\texttt{yaoyan@stu.cqnu.edu.cn}\\[0.5em]
\small College of Computer and Information Science; Chongqing Key Lab of Cognitive Intelligence and Intelligent Finance
}

\begin{document}
\maketitle

\blfootnote{Accepted to EMNLP 2025. }
\begin{abstract}
Arithmetic offers a compact test of whether large language models compute or memorize. We study multi-digit addition in \textsc{LLaMA-3-8B-Instruct} using linear probes and the logit lens, and find a consistent four-stage, layer-wise ordering of probe-decodable signal types across depth: (1) early layers encode formula structure (operand/operator layout) while the gold next token is still far from top-1; (2) mid layers expose digit-wise sums and carry indicators; (3) deeper layers express result-level numerical abstractions that support near-perfect digit decoding from hidden states; and (4) near the output, representations align with final sequence generation, with the correct next token reliably ranked first. Across experiments, each signal family becomes linearly decodable with high accuracy (stage-wise peaks typically $\geq$95\% on in-domain multi-digit addition, and up to 99\%). Taken together, these observations—\emph{in our setting}—are consistent with a hierarchical, computation-first account rather than rote pattern matching, and help explain why logit lens inspection is most informative mainly in later layers.Code and data are available at\\
\url{https://github.com/YaoToolChest/addition-in-four-movements.git}.
\end{abstract}

\section{Introduction}
\noindent\textbf{Status.} This paper has been accepted to EMNLP 2025.
Generalizable reasoning remains a core challenge for large language models (LLMs). Arithmetic-and multi-digit addition in particular-offers a controlled probe of whether models execute algorithmic procedures or merely match surface patterns \cite{nikankin2024arithmeticalgorithmslanguagemodels}. Humans solve addition via a well-defined carry algorithm; an LLM could either approximate this stepwise computation (e.g., column-wise summation with carry propagation) or rely on shortcuts. Consistent, systematic performance across inputs is suggestive of internal computation, whereas systematic or context-specific failures point to superficial heuristics.

Prior work advances two contrasting views. One line of evidence reports structured, algorithm-like behavior: analyses of GPT-style models reveal components that specialize in partial summation and carry handling \cite{yu-ananiadou-2024-interpreting}, and even minimal two-layer encoder-only Transformers can learn the carry-addition procedure \cite{kruthoff2024carryingalgorithmtransformers}. Another line argues that correct answers often arise from simple heuristics \cite{nikankin2024arithmeticalgorithmslanguagemodels}, citing a ``bag of heuristics'' in neuron activations \cite{NEURIPS2024_2cc8dc30} and out-of-distribution errors such as commutativity violations or symbol perturbations \cite{yan2025phdlevelllmstrulygrasp}. Related analyses report Fourier-like representational structure that mixes low- and high-frequency features \cite{NEURIPS2024_2cc8dc30}. These disagreements leave open how mainstream LLMs implement addition.

\begin{figure}[t!]
\includegraphics[width=\columnwidth]{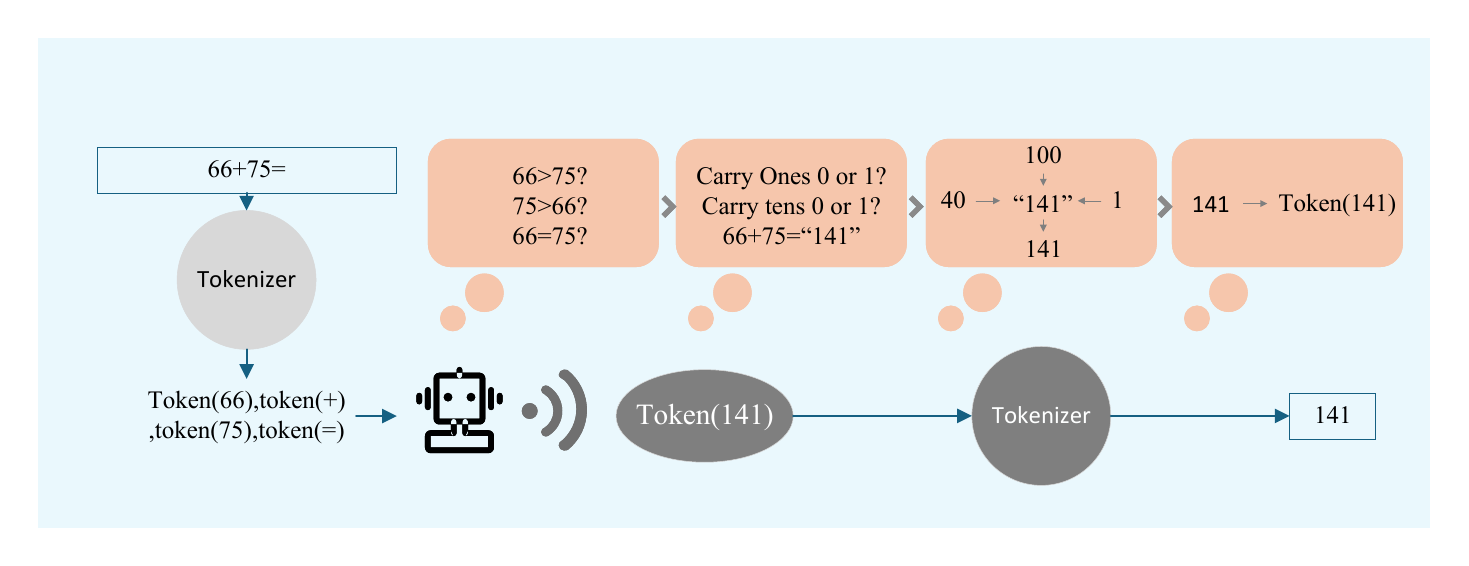}
\caption{Multi-stage information flow for addition in LLMs. As depth increases, internal representations sequentially expose (1) problem structure (operand/operator layout), (2) core computations (column sums and carries), (3) result-level numerical abstractions (digit identities), and (4) output-aligned organization that drives final token generation. Each signal type is \emph{operationalized} via a linear-probe task and becomes decodable at different depths.}
\label{fig:addition-order}
\end{figure}

We focus on LLaMA-3-8B-Instruct \cite{grattafiori2024llama} and ask: does it perform multi-digit addition through structured, algorithmic computation (e.g., digit-wise summation with carry propagation), or through non-algorithmic pattern extraction?

Our starting point is a logit lens analysis \cite{yu-ananiadou-2024-interpreting}. Consistent with observations that the lens is most informative in later layers (cf. \cite{DBLP:conf/emnlp/KatzBGW24}), we find that next-token probabilities largely reflect compressed late-stage representations and reveal little about intermediate processing. Because the logit lens is informative mainly in late layers-i.e., the gold next token tends to become rank-1 only late in the forward pass-we hypothesize a hierarchical, multi-stage computation rather than a simple accumulation of confidence on the correct token. In other words, intermediate layers appear to reorganize information before output selection, rather than merely amplifying the correct-token probability.

Guided by this observation, we articulate-without presupposing discrete stage boundaries or causal transitions-a layer-wise organization of addition-related signals that become decodable in a consistent order as depth increases. We make each signal family explicit and testable:

\begin{enumerate}
\item \textbf{Formula-structure signals}: encoding of operand/operator layout and delimiter positions; probed by predicting token-category labels and structural spans from hidden states.
\item \textbf{Core-computation signals}: digit-wise sums and carry indicators; probed by classifiers for per-column sum mod~10 and carry-bit presence.
\item \textbf{Numerical-abstraction signals}: result-level representations enabling accurate digit identity decoding independent of local context; probed by position-conditioned digit decoders and cross-position generalization tests.
\item \textbf{Output-organization/generation signals}: alignment of representations with sequence generation that yields reliable top-1 ranking of the correct next token; assessed by logit lens agreement and calibrated rank metrics.
\end{enumerate}

Using linear probes on LLaMA-3-8B-Instruct, we observe that these four signal families emerge in the above order across layers. The ordering supports a computation-first interpretation in our setting, reconciles component specialization with reusable numerical structure, and explains why logit lens inspection is most revealing in later layers. We emphasize that our analysis is diagnostic rather than interventional: decodability does not imply necessity, and we do not claim causal stage transitions; instead, we provide operational evidence for a structured information-processing pipeline during addition.

\section{Related Work}\label{sec:related}

We organize prior work on LLM arithmetic competence into three lenses: (i) algorithmic computation, (ii) heuristic pattern matching, and (iii) abstract numerical representation.

\paragraph{Algorithmic evidence}
Toy models and mid-sized LLMs exhibit algorithm-like solutions: attention can implement DFT-style modular addition \cite{DBLP:conf/iclr/NandaCLSS23}, and helical/``clock'' number encodings have been proposed for general addition \cite{kantamneni2025language}. In larger models, component specialization persists --- carry-routing heads and layer-wise decompositions into column summation and carry propagation are reported in 7B-scale GPT/LLaMA variants \cite{yu-ananiadou-2024-interpreting,kruthoff2024carryingalgorithmtransformers}; causal analyses further identify operand routers and parallel digit streams \cite{DBLP:conf/emnlp/StolfoBS23,DBLP:conf/iclr/QuirkeB24}.

\paragraph{Heuristic evidence}
An opposing view attributes success to a ``bag of heuristics'': neuron-level studies find sparse, template-like strategies \cite{nikankin2024arithmeticalgorithmslanguagemodels}, and controlled out-of-distribution tests reveal failures under digit permutations, novel symbols, or simple reorderings \cite{yan2025phdlevelllmstrulygrasp}.

\paragraph{Abstract numerical representations}
Beyond this dichotomy, pre-trained LLMs appear to superimpose magnitude and parity in a Fourier-like basis \cite{NEURIPS2024_2cc8dc30}; digit values are linearly recoverable, suggesting a low-dimensional numerical subspace \cite{DBLP:conf/coling/ZhuDS25}. Group-structured arithmetic elicits basis vectors reminiscent of representation theory \cite{chughtai2023neural}, tree-structured reasoning traces can be reconstructed from hidden states \cite{DBLP:conf/emnlp/HouLFSZZBS23}, and evidence supports digit-by-digit encoding with position-specific neuron groups \cite{levy-geva-2025-language}.

\paragraph{Our position}
We provide a layer-resolved account of multi-digit addition in LLaMA-3-8B by combining linear probes \cite{DBLP:journals/coling/Belinkov22} with the logit lens \cite{nostalgebraist2020logitlens}. The observed four-stage trajectory-from formula recognition, through core computation, to numerical abstraction and output generation-bridges component specialization with the emergence of reusable numerical structure.

\begin{figure*}[t]
  \centering
  \includegraphics[width=\linewidth]{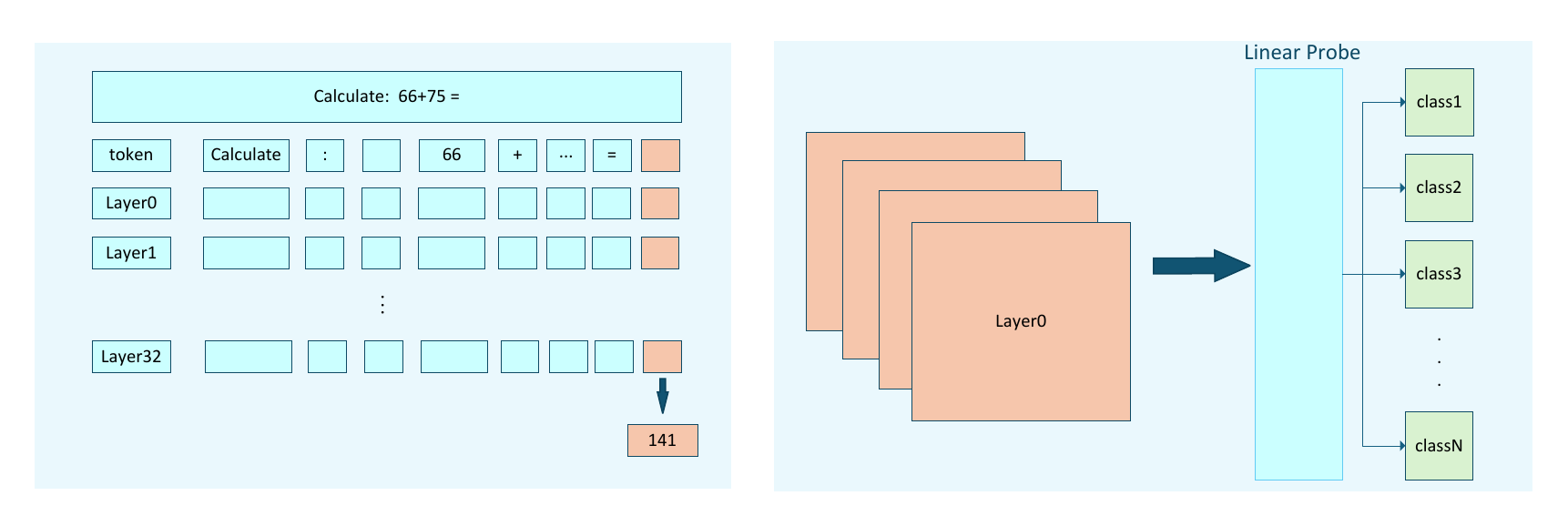}
  \caption{Representation extraction sites and diagnostics for arithmetic reasoning. For an addition prompt (\texttt{``Calculate: num1 + num2 = ''}),we record the hidden state of the last input token at every \textbf{layer state} (L0--L32) of \textsc{LLaMA-3-8B-Instruct}. At layer $l$, we compute (i) a logit lens projection of the last-token activation, $\boldsymbol{\ell}^{(l)}=\mathbf{h}_S^{(l)} W_U^{\top}$, which gives the next-token logits just before the first answer token is generated; and (ii) a linear probe over the same vector, $\mathbf{o}^{(l)}=W^{(l)}\mathbf{h}_S^{(l)}+b^{(l)}$, trained to decode task-relevant attributes (e.g., carry bits, digit values). Scanning $l$ localizes where arithmetic information becomes linearly accessible and visualizes how predictions sharpen across depth.}
  \label{fig:extraction}
\end{figure*}

\section{Methodology}

\paragraph{Overview.}
We study how internal representations evolve when \textsc{LLaMA-3-8B-Instruct} performs multi-digit addition. As illustrated in Fig.~\ref{fig:extraction}, for each input we extract the hidden state of the last input token (the position that predicts the first answer token) at every layer and apply two complementary diagnostics: (i) \emph{logit lens} to observe the model’s layer-wise next-token distribution, and (ii) \emph{linear probes} to test whether specific arithmetic attributes are linearly decodable. This setup allows us to map where and when structure, intermediate computations, and result digits emerge across depth.
\paragraph{Indexing convention.}
We adopt a \textbf{0-indexed} convention with \textbf{layer states} \(l \in \{0,\dots,32\}\) for \textsc{LLaMA-3-8B-Instruct}.
Here, \(l{=}0\) denotes the \textbf{embedding state} (before any Transformer block), and \(l{=}1,\dots,32\) denote the outputs \textbf{after} Transformer blocks 1-32.
Unless otherwise noted, all results are reported using layer states \(L0\)-\(L32\).

\subsection{Model, Prompting, and Target Activations}
We use a standardized prompt template \texttt{``Calculate: num1+num2 = ''} (note the trailing space), which tokenizes to $X=(x_1,\dots,x_S)$. The Transformer yields a sequence of hidden states $\mathbf{H}^{(l)}\in\mathbb{R}^{S\times d}$ at each \textbf{layer state} $l\in\{0,\dots,32\}$, with the last-token vector denoted $\mathbf{h}_S^{(l)}\in\mathbb{R}^{d}$. We focus on $\mathbf{h}_S^{(l)}$ because it conditions the generation of the first answer token. Among several open-source LLMs we screened, \textsc{LLaMA-3-8B-Instruct} offered the most pronounced and consistent layer-wise probe signals and stable performance on our addition benchmark (Table~\ref{tab:llm_arithmetic_comparison}), so we use it as our primary subject. Results for other models, including math-optimized variants (Mistral-7B-Instruct~\cite{jiang2023mistral7b}, Qwen2.5-Math-7B-Instruct~\cite{DBLP:journals/corr/abs-2409-12122}, AceMath-7B-Instruct~\cite{DBLP:conf/acl/LiuCSCP25}), are reported in Appendix~\ref{sec:model_experiments}.

\subsection{Logit Lens}
Given the model’s unembedding matrix $W_U\in\mathbb{R}^{|\mathcal{V}|\times d}$, the logit lens projects the last-token state at layer $l$ to layer-specific next-token logits
\[
\boldsymbol{\ell}^{(l)}=\mathbf{h}_S^{(l)} W_U^{\top}\in\mathbb{R}^{|\mathcal{V}|}.
\]
Interpreting $\mathrm{softmax}(\boldsymbol{\ell}^{(l)})$ as the distribution the model would produce if layer $l$ were final, we can track how the rank of the gold next token and the sharpness of the distribution evolve with depth.

\begin{table}[t!]
\centering
\begin{tabular}{lc}
\hline
\textbf{Model} & \textbf{Addition Accuracy} \\
\hline
\textsc{LLaMA-3-8B-Instruct} & 98.18\% \\
Mistral-7B-Instruct & 96.12\% \\
AceMath$^{\dagger}$ & 99.10\% \\
Qwen2.5-Math-7B$^{\dagger}$ & 98.68\% \\
\hline
\end{tabular}
\caption{Overall accuracy on 1–6 digit addition (exact match). $^{\dagger}$ denotes math-optimized models. Although a math-optimized model attains the highest accuracy, \textsc{LLaMA-3-8B-Instruct} exhibits the clearest and most consistent layer-wise probe signals (Sections~\ref{sec:experiments}--\ref{sec:discussion}), so we focus on it for the mechanistic analyses. See Appendix~\ref{sec:Metrics} for evaluation details and Appendix~\ref{sec:model_experiments} for per-model probe curves and results on the other models.}
\label{tab:llm_arithmetic_comparison}
\end{table}

\subsection{Linear Probes: Design, Training, and Evaluation}
To assess where arithmetic information becomes linearly accessible, we train a separate linear classifier at each layer $l$ for each attribute (e.g., carry bits at each position, per-digit sums, result digits). Each probe takes the last-token vector $\mathbf{h}_S^{(l)}\in\mathbb{R}^{d}$ as input and produces $K$ logits:
\begin{align}
\mathbf{o}^{(l)} &= W^{(l)} \mathbf{h}_S^{(l)} + b^{(l)}, \label{eq:probe}\\
\hat{\mathbf{p}}^{(l)} &= \mathrm{softmax}\!\big(\mathbf{o}^{(l)}\big). \label{eq:softmax}
\end{align}
Here $W^{(l)}\!\in\!\mathbb{R}^{K\times d}$ and $b^{(l)}\!\in\!\mathbb{R}^{K}$. We optimize the standard cross-entropy over a training set of size $N$:
\begin{equation}
\begin{aligned}
\mathcal{L}^{(l)} &= -\frac{1}{N}\sum_{i=1}^{N}\log \hat{p}^{(l)}_{i,y_i} \\
&= -\frac{1}{N}\sum_{i=1}^{N}\sum_{k=1}^{K}\mathbb{1}[y_i=k]\log \hat{p}^{(l)}_{i,k}.
\end{aligned}
\label{eq:xent}
\end{equation}
and report accuracy on a held-out test set.\footnote{Unless otherwise noted, the base model is frozen; probes are trained for 10 epochs and the checkpoint is selected by validation accuracy. Implementation details are in Appendix~\ref{sec:AppGeneral}.}
Layer-wise accuracy curves reveal where each attribute becomes reliably decodable.

\subsection{Data Handling and Evaluation Protocol}
\textbf{Dataset construction.}
We construct task-specific datasets for probe training and testing using the above prompt template (with minor variants when stated). Splits are disjoint at the instance level, near-duplicates are removed, and OOD sets (e.g., unseen operand ranges/formats) are included when applicable.

\noindent\textbf{Preprocessing and sampling.}
We standardize spacing and punctuation, control operand distributions (e.g., uniform within range), balance classes where relevant, check tokenization artifacts, and fix random seeds for reproducibility (Appendix~\ref{sec:AppGeneral}).

\noindent\textbf{Evaluation criteria.}
For each attribute and layer, we compute accuracy on held-out data. When accuracy is near-perfect and stable across seeds, we treat the corresponding representation as \emph{linearly accessible} at that layer. Comparing the onset of high decodability across attributes yields a layer-wise ordering that characterizes the flow of arithmetic information.

\paragraph{Model selection rationale.}
Our goal is to diagnose \emph{how} models compute, rather than to optimize leaderboard accuracy. We therefore prioritize a widely used, instruction-tuned baseline whose internal signals are clear and reproducible under linear probes. While math-optimized models (marked $^{\dagger}$ in Table~\ref{tab:llm_arithmetic_comparison}) can achieve slightly higher raw accuracy on our benchmark, their specialized training can introduce confounds for mechanistic attribution. In contrast, \textsc{LLaMA-3-8B-Instruct} consistently exhibits a clean four-stage ordering across layers in our probes, making it a suitable primary case study. Appendix~\ref{sec:model_experiments} shows that the same ordering largely persists across other models with minor depth shifts.

\section{Experiments and Results}\label{sec:experiments}
We conduct a series of experiments to diagnose how arithmetic features are represented within the \textsc{LLaMA-3-8B-Instruct} model.
Across all experiments, diagnostic probes are trained on the hidden states of \textbf{all layer states} L0--L32 (embedding + 32 Transformer blocks), while the pretrained model parameters remain frozen.
We use a common default setup for datasets, prompts, preprocessing, and splits; experiment-specific settings are described in each subsection.

\subsection{Formula structure representation}

\begin{figure}[htbp!]
\includegraphics[width=\columnwidth]{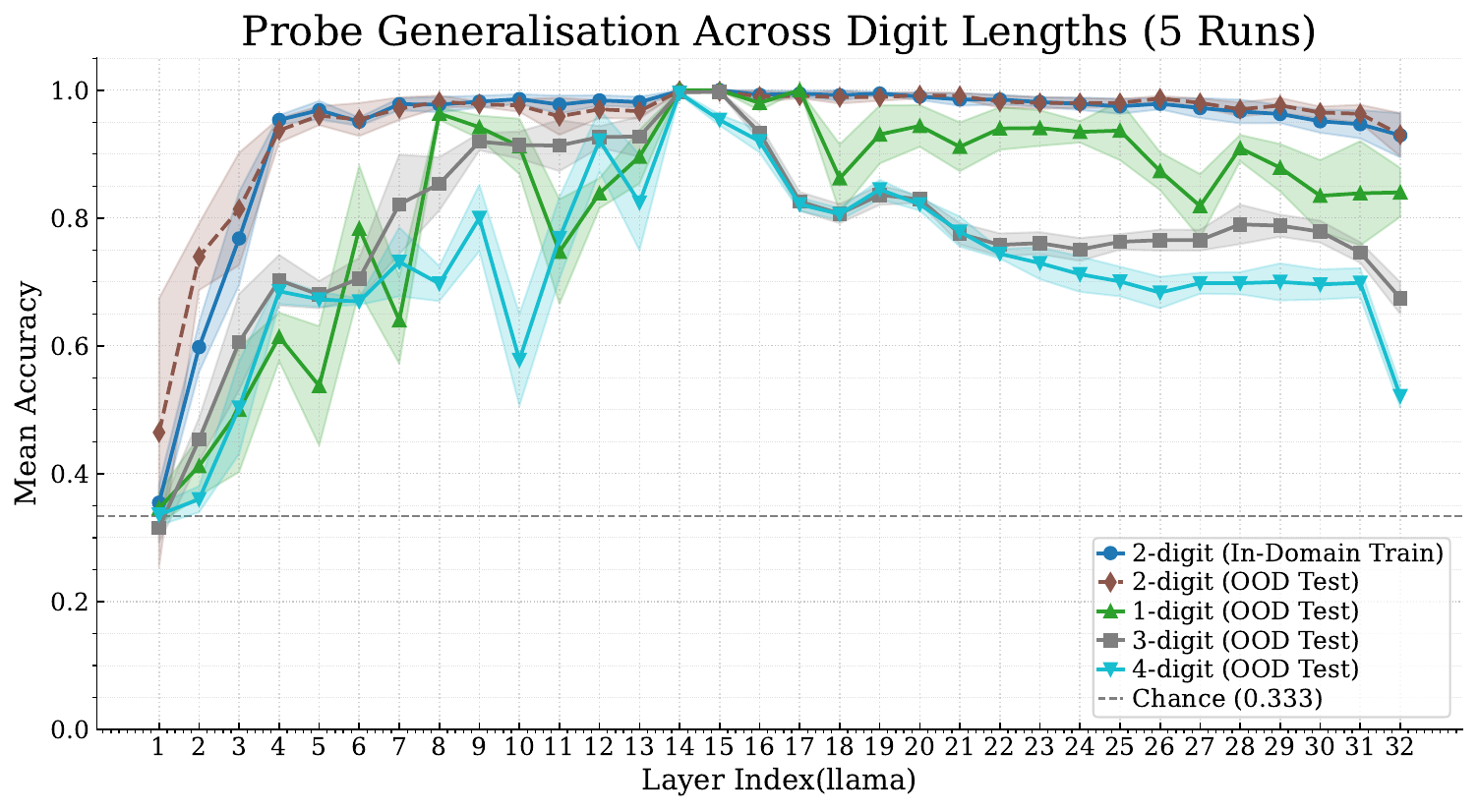}
\caption{\textbf{Layer-wise decodability of formula structure} (3-way classification: $a{+}b$, $b{+}a$, $a{+}a$ with $a>b$).
At each \textbf{layer state} (L0--L32) of \textsc{LLaMA-3-8B-Instruct}, we train a linear probe on the last-input-token state $\mathbf{h}_S^{(l)}$ to predict the equation type.
Probes are trained on two-digit expressions (in-domain) and evaluated on a disjoint test split comprising the in-domain two-digit set and out-of-distribution (OOD) lengths (1-, 3-, and 4-digit).
Curves show mean accuracy over five independent runs; the dashed line marks the random baseline ($1/3$).
Accuracy rises from near-baseline at shallow layers to stable, high values in early-to-mid layers, indicating that structural cues about operand order and self-addition become increasingly linearly readable with depth.}
\label{fig:Addition_Order_Experiments}
\end{figure}

\paragraph{Arithmetic Structure Recognition Experiments:}
We ask whether the model encodes the \emph{structure} of an addition prompt-independently of the specific numbers-early in the forward pass.
For each layer $l$, we extract the last-input-token vector $\mathbf{h}_S^{(l)}$ and fit a 3-way linear classifier to distinguish $\{a{+}b,\,b{+}a,\,a{+}a\}$ (with $a>b$).
Training uses two-digit expressions; evaluation covers a held-out in-domain two-digit set and out-of-distribution (OOD) lengths (1-, 3-, and 4-digit).
As shown in Fig.~\ref{fig:Addition_Order_Experiments}, performance starts near the $1/3$ random baseline at shallow layers, increases with depth, and plateaus at high accuracy in mid-to-late layers.
The narrowing gap between OOD and in-domain curves suggests that the network progressively abstracts operand-order and self-addition patterns into linearly accessible features before answer generation.

\subsection{Emergence of core computational features}

\paragraph{Sum-range classification (``arithmetic results'' probes).}
We localize where the \emph{sum} becomes linearly decodable by training layer-wise probes to classify problems into contiguous sum bins (e.g., $500$-$509$), which yields a 10-way task with a $0.10$ random baseline. Each class contains 400 samples; dataset generation and splits are detailed in Appendix~\ref{sec:app-exp-details}. As shown in Fig.~\ref{fig:sum_range_probe}, accuracy is near baseline in shallow layers (L0-L8), increases gradually from $\sim$L8, exhibits a sharp rise around L16-L19, and then plateaus at high accuracy from about L19 onward, indicating that aggregate result information consolidates in mid-to-late layers.

\begin{figure}[htbp!]
\includegraphics[width=\columnwidth]{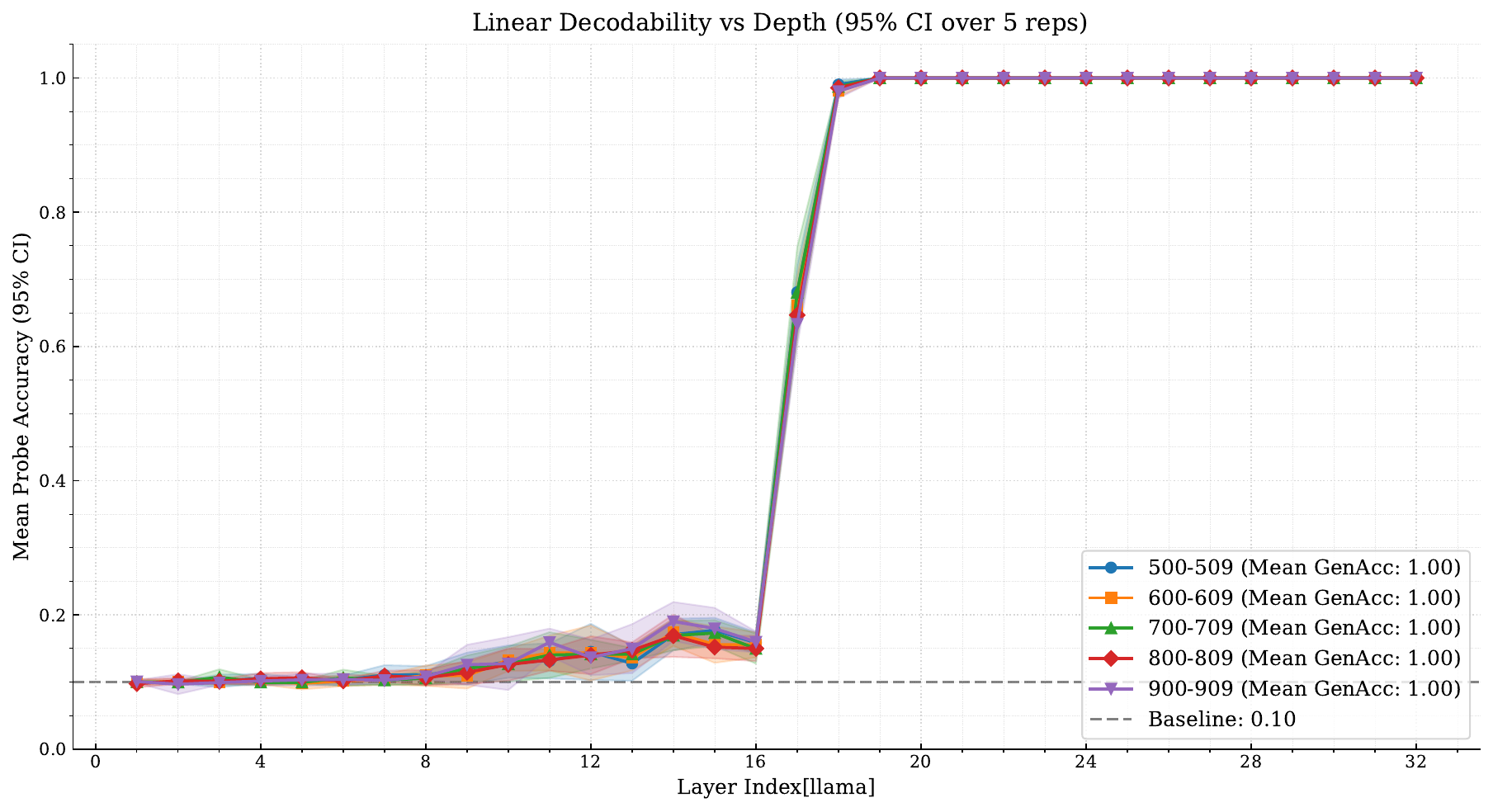}
\caption{\textbf{Sum-range decodability across depth.} Layer-wise linear probes classify addition problems into contiguous 10-value sum bins (e.g., $500$-$509$, $900$-$909$). All ranges show a common profile: near-baseline performance in early layers, a marked transition around L16-L19, and a stable high-accuracy plateau thereafter (random baseline $=0.10$).}
\label{fig:sum_range_probe}
\end{figure}

\begin{figure}[htbp!]
\includegraphics[width=\columnwidth]{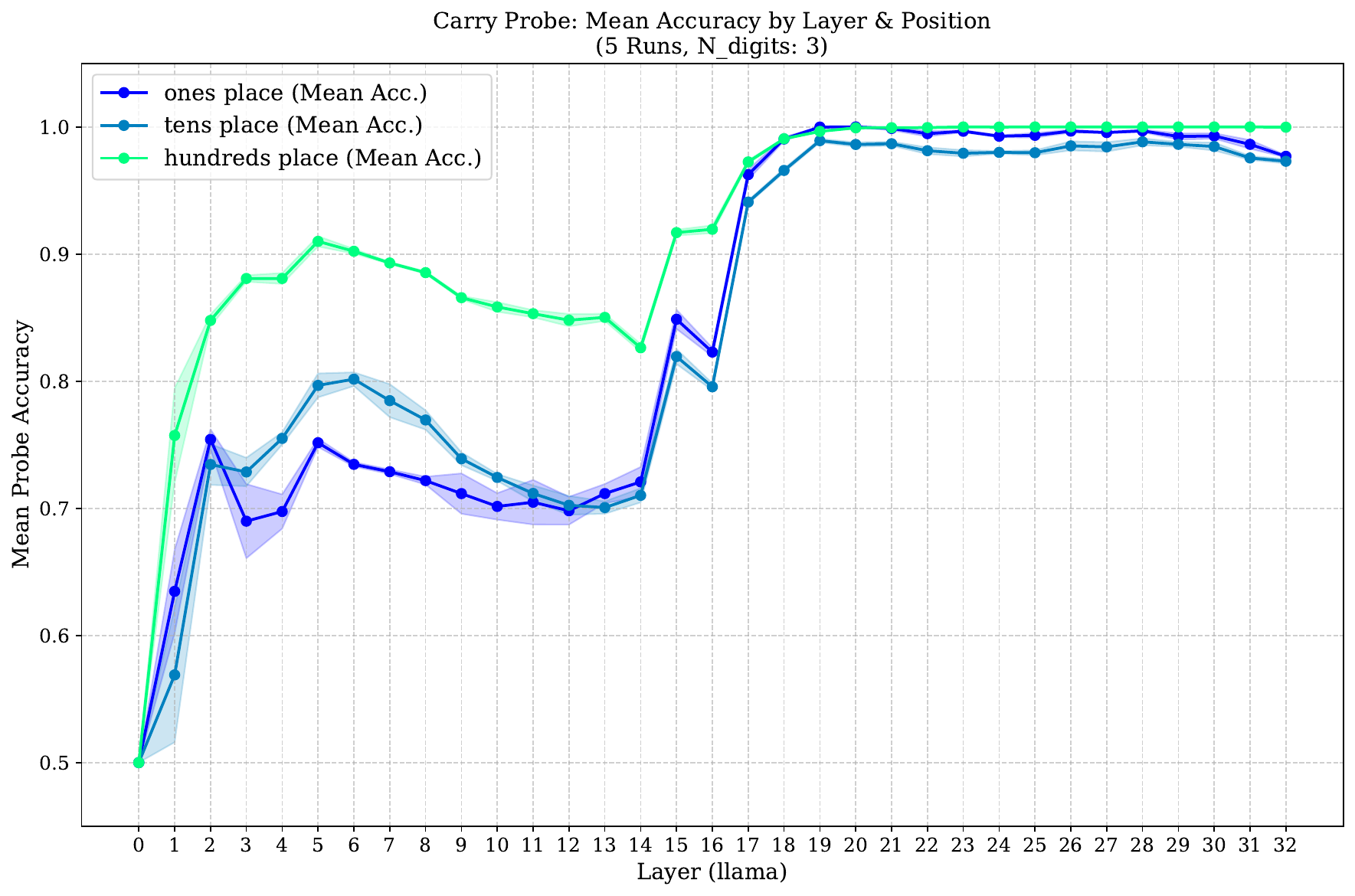}
\caption{\textbf{Carry detection by position (3-digit addition).} Mean probe accuracy by layer for ones/tens/hundreds positions (5 runs; balanced labels). Curves rise from the $0.50$ random baseline, begin increasing consistently around $\sim$L14, and approach ceiling by $\sim$L19, indicating efficient carry processing in mid-to-late layers.}
\label{fig:Carry_signal_experiment}
\end{figure}

\paragraph{Carry-signal detection.}
To test when carry information emerges, we train \emph{independent} binary probes (balanced carry/no-carry) for the ones, tens, and hundreds positions on three-digit addition. Accuracy rises from near the $0.50$ random baseline in early layers, begins a steady climb around $\sim$L14, and approaches saturation for all positions by $\sim$L19 (Fig.~\ref{fig:Carry_signal_experiment}). This alignment with the sum-range transition supports the view that carry computation is established in mid layers and consolidated before output formation.

\subsection{Numerical abstraction of results}

\paragraph{Digit-wise decoding (numerical abstraction).}
We probe when individual \emph{result digits} become linearly accessible by training separate 10-way (digits $0$-$9$) linear classifiers at each layer on the last-input-token state $\mathbf{h}_S^{(l)}$, one probe per position (ones, tens, hundreds). This yields a random baseline of $0.10$ per task; dataset construction and splits are detailed in Appendix~\ref{sec:app-digit-id}. As shown in Fig.~\ref{fig:digit_abstraction}, accuracy rises from near-baseline in shallow layers, improves steadily from mid layers, and reaches a stable high plateau in late layers (e.g., after $\sim$L28). The three positions exhibit similar depth profiles, indicating that digit-level abstractions consolidate late in the forward pass.

\begin{figure}[htbp!]
\includegraphics[width=\columnwidth]{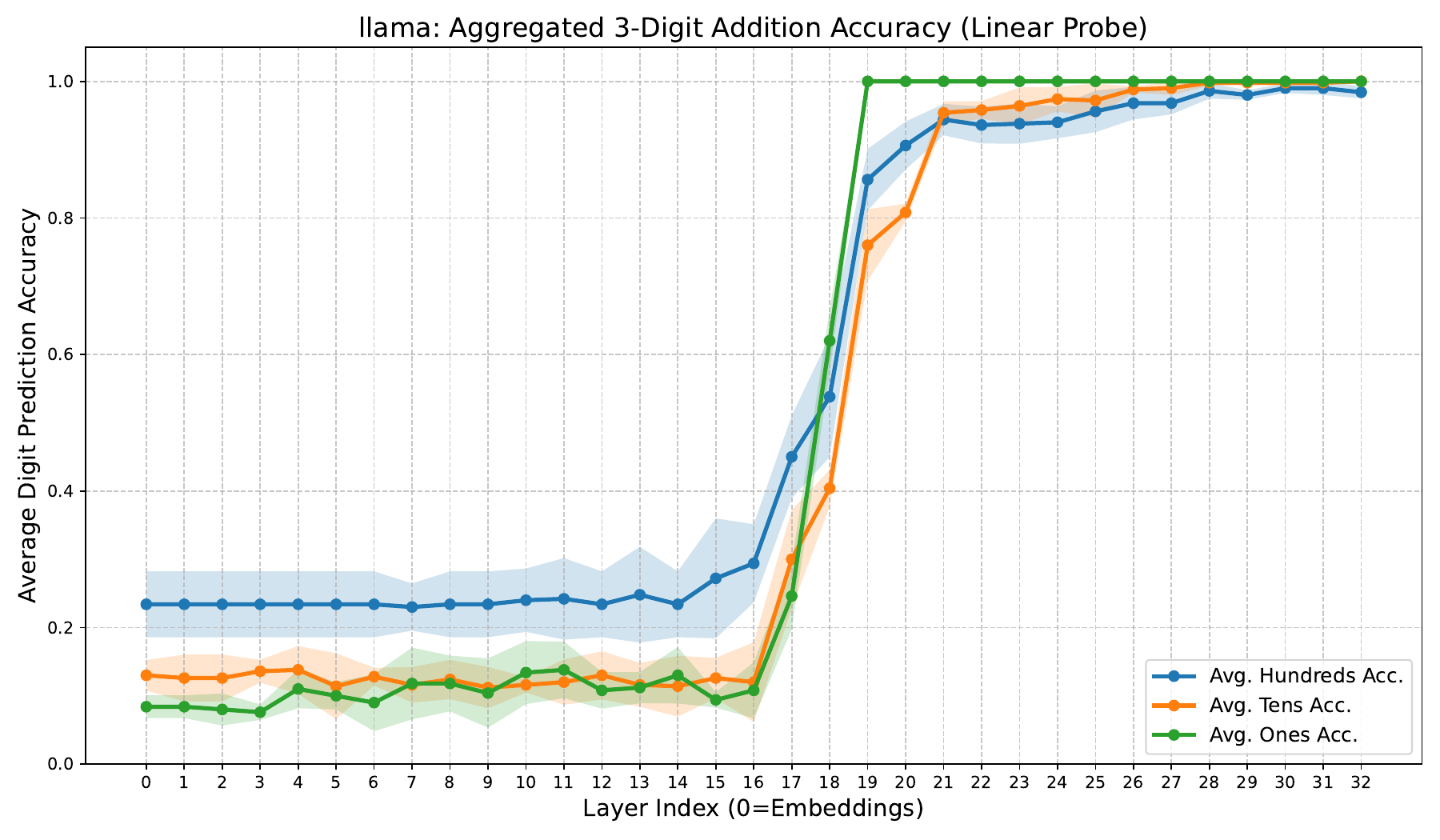}
\caption{\textbf{Digit-wise decodability of 3-digit sums.} Mean probe accuracy ($\pm$95\% CI over 5 runs) by layer for predicting the ones, tens, and hundreds digits from $\mathbf{h}_S^{(l)}$. All three tasks improve with depth from near the 10-class chance level ($0.10$) and saturate at high accuracy in late layers (e.g., after $\sim$L28), indicating robust numerical abstraction across positions.}
\label{fig:digit_abstraction}
\end{figure}

\begin{figure}[htbp!]
\includegraphics[width=\columnwidth]{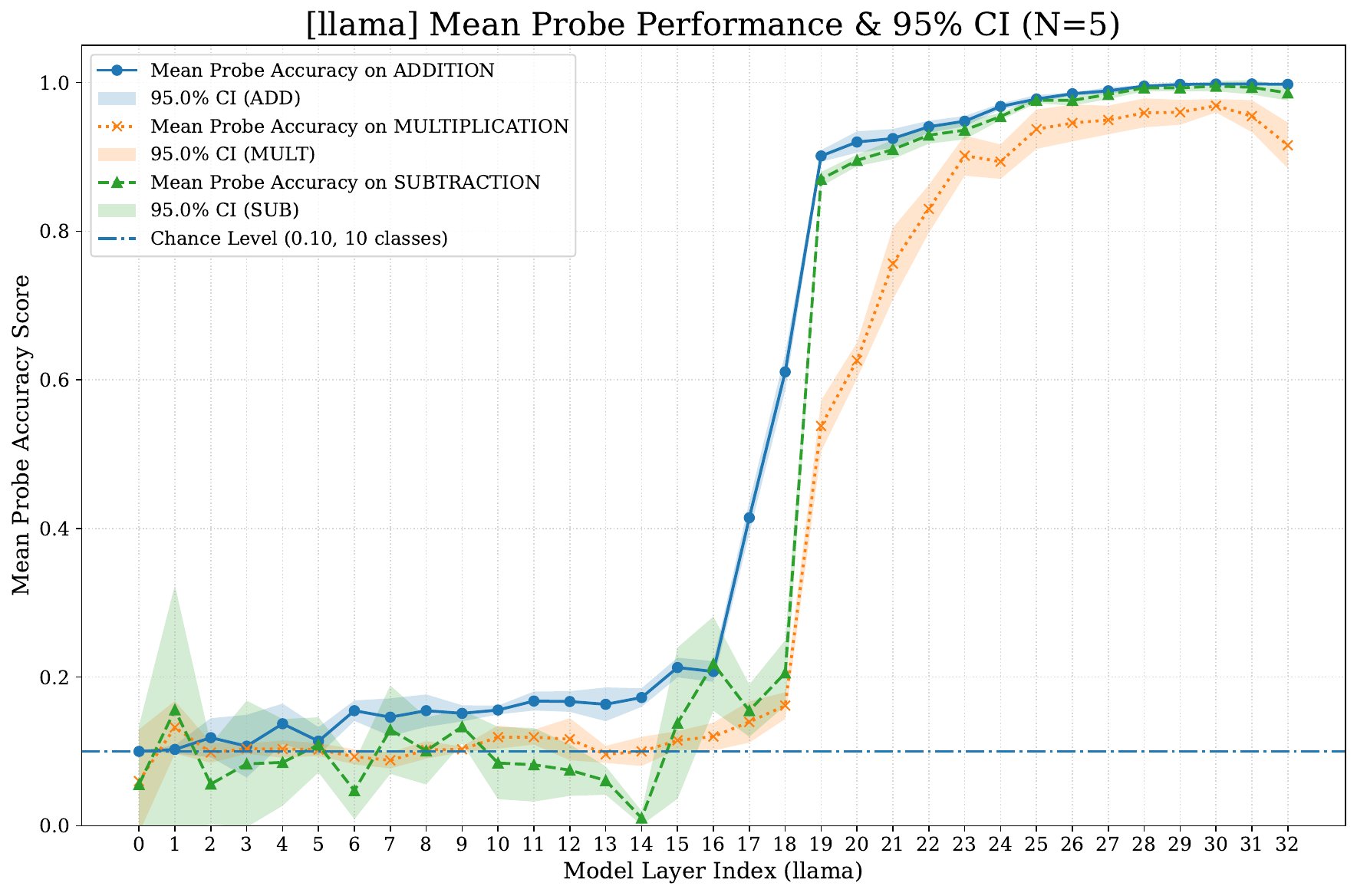}
\caption{\textbf{Hundreds-digit probe generalization across operations.} Mean accuracy ($\pm$95\% CI over 5 runs) by layer when a probe trained on \emph{addition} is tested on \emph{addition}, \emph{subtraction}, and \emph{multiplication}. All curves exceed the 10-class chance level ($0.10$); subtraction generalizes more strongly than multiplication.}
\label{fig:hundreds_generalization}
\end{figure}

\paragraph{Cross-operation generalization.}
To assess whether these digit representations transfer across operations, we train the hundreds-digit probe on \emph{addition} and evaluate it on held-out \emph{subtraction} and \emph{multiplication}. Figure~\ref{fig:hundreds_generalization} shows that generalization is strongest to subtraction (peaking near $\sim$0.9) and somewhat lower to multiplication (peaking near $\sim$0.8), with both tasks well above the 10-class chance level ($0.10$). This pattern suggests a shared numerical substrate with operation-specific components that are less aligned for multiplication than subtraction. 

\subsection{Logit lens: Organization and generation of output content}

\paragraph{First-correct-token emergence (logit lens).}
We use the logit lens to identify the earliest layer at which the gold \emph{next token} (the first digit of the sum) becomes the model’s top-1 prediction.
For each layer $l$, we project the last-input-token state $\mathbf{h}_S^{(l)}$ through the unembedding matrix to obtain logits and record the first $l$ where the gold token attains rank~1 (The dataset consists of addition problems where two 3-digit addends result in a 3-digit sum; details in Appendix~\ref{sec:app_logit_lens}).
As shown in Fig.~\ref{fig:logit_lens_first_correct_token}, all test prompts first reach top-1 within late \textbf{layer states} (L23-L32), with a pronounced peak at L30 (about $45.4\%$ of samples across five runs) and no “never top-1” cases.
This indicates that decisive arithmetic commitment and output selection are finalized in the concluding third of the network, consistent with mid-layer transitions seen in sum-range and carry probes but occurring later in the computation pipeline.

\begin{figure}[htbp!]
\includegraphics[width=\columnwidth]{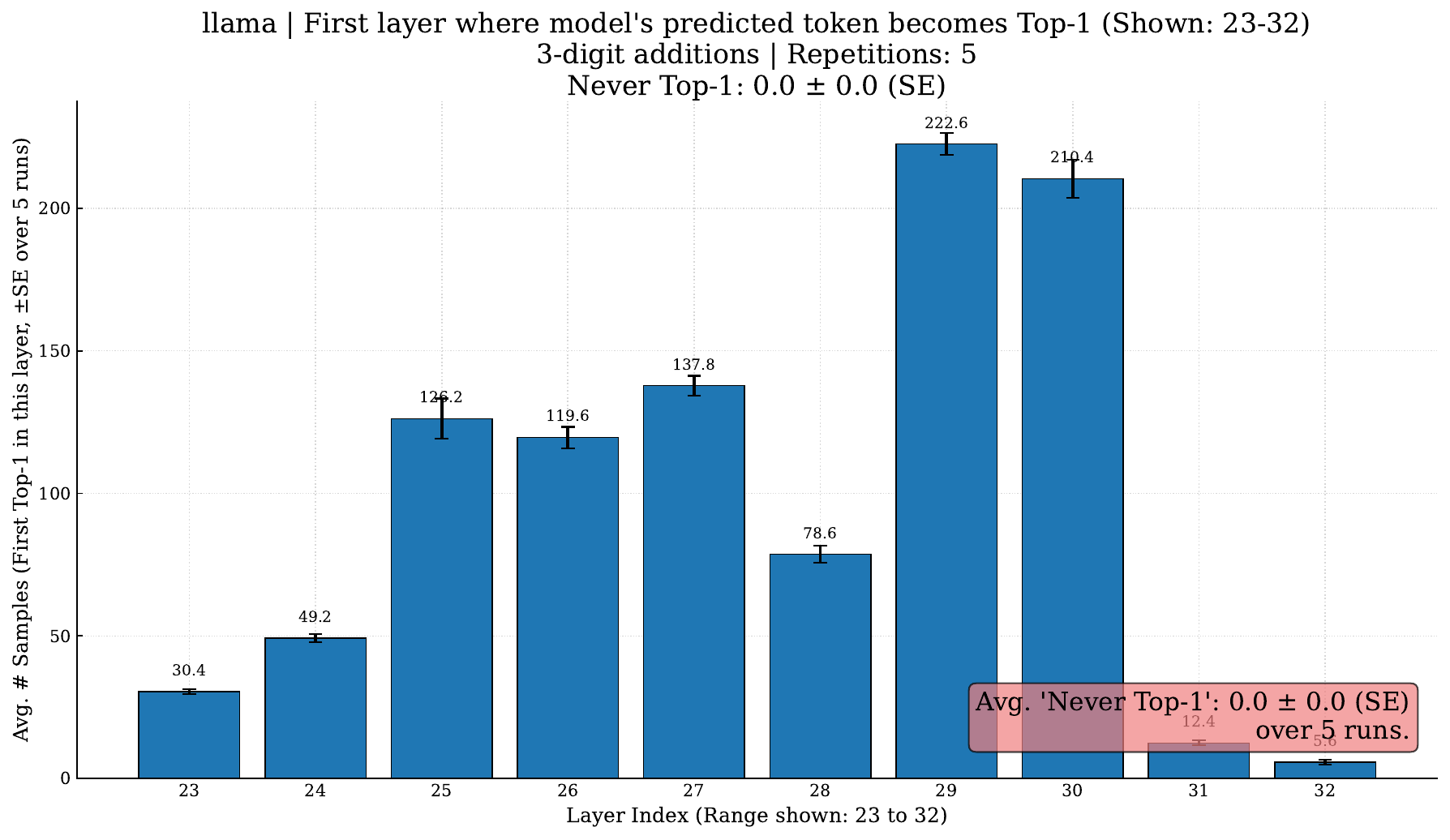}
\caption{\textbf{Earliest layer where the gold next token becomes top-1 (logit lens).}
Histogram over $N{=}1000$ three-digit additions showing, for each \textbf{layer state} in L23--L32, the mean number of samples whose gold token \emph{first} becomes top-1.
All examples first reach top-1 in this late-layer range, peaking at L30 ($\sim$45.4\% of samples).}
\label{fig:logit_lens_first_correct_token}
\end{figure}

\section{Analysis}

The experimental results of this study systematically reveal that when the \textsc{LLaMA-3-8B-Instruct} large language model processes addition operations, the evolution of its internal information representation presents a clear, sequential trajectory.

Firstly, in the shallower layers of the network (e.g., in the formula structure recognition experiment, probes trained on two-digit numbers, when processing four-digit comparisons, show peak accuracy around layer 14), the model can efficiently identify the structural type of addition expressions (such as a+b, b+a, a+a). At this point, the probe accuracy for decoding these structural features is high, especially its good generalization to different operand lengths, indicating that the model can capture the syntactic and basic semantic composition of the input arithmetic task at early computational depths.

Secondly, in the middle layers of the network (roughly from layer 14 to layer 18), the decodability of core arithmetic operation features begins to become prominent. The decoding accuracy for the representation of carry signals (whether for units, tens, or hundreds digits) significantly improves in this interval, reaching a near-perfect level around layer 19, forming a clear and easily decodable representation of the key steps of the operation. Almost simultaneously (starting with a sharp rise from layer 16 and stabilizing around layers 17-19), the prediction accuracy for the overall addition result (i.e., the "sum") also reaches saturation, suggesting that a stable internal representation of the result is largely formed by these layers.

Subsequently, in the deeper layers of the network (especially after layer 25), probe results indicate that the model appears to focus more on a detailed numerical abstraction of the operation result. Experiments show that in these deeper layers, the model can accurately distinguish and represent the individual digits composing the "sum" (such as the hundreds digit, tens digit, units digit), linking the operation result with its specific numerical attributes. Logit Lens analysis further indicates that the generation of the final answer token primarily occurs in the final stages of the network (layers 23 to 32, peaking at layer 30). This is sequentially consistent with the maturation point of internal numerical representations (after layer 25), suggesting a potential transformation process within the model from internal numerical concepts to the final text output.

Furthermore, in the "arithmetic result experiment," the probe accuracy for sums across different numerical ranges (500-509 to 900-909) shows a highly consistent trend with increasing layer depth, all reaching high accuracy at similar levels (around layers 16-19). This indicates that, from the perspective of the evolution pattern of probe accuracy, these specific numerical range problems do not exhibit a significant difficulty gradient in terms of "when they become solvable" within the model's internal processing.

\section{Discussion}\label{sec:discussion}

The sequential evolution pattern of information representation observed in this experiment provides important clues for understanding how LLMs perform arithmetic operations. Several phenomena are worth further discussion:

Firstly, \textbf{representations are dynamic}, with their prominence peaking at specific positions in the processing trajectory, sometimes followed by a slight decrease in decodability. This 'low-high-slight decrease' pattern was observed in several probe experiments. For instance, in the formula structure representation task (Fig.~\ref{fig:Addition_Order_Experiments}, focusing on probes trained on 2-digit addition), the accuracy for identifying '2-digit (In-Domain Train)' patterns peaks around layer 10 and then shows a gradual decline. Similarly, the '4-digit (OOD Test)' curve in the same figure peaks around layer 14 before decreasing. A similar trend is observable in the carry signal experiment (Fig.~\ref{fig:Carry_signal_experiment}). For example, the 'hundreds place' carry detection accuracy sharply peaks around layer 19-20, then exhibits a noticeable decrease in subsequent layers before a slight rebound. The 'tens place' and 'ones place' also show peaks (around layer 19 and 18 respectively) followed by a dip.

This pattern suggests that initially, accuracy is low as the specific arithmetic representation has not yet clearly formed. In intermediate layers, this information becomes most linearly decodable, and its representation strength and accessibility reach their peak. Subsequently, as the model processes this information or integrates it for subsequent computational stages, it may shift focus. Newly accumulated representational information from later layers (via residual connections) or the transformation of these features for downstream tasks might slightly interfere with or partially obscure the previously formed representations that are no longer central to the immediate next step of computation. This reflects the dynamic evolutionary nature of the model's internal representations, where specific information (as detected by our probes) is most clearly and accessibly represented at particular depths, rather than being monotonically refined.

Secondly,\textbf{Generalizability of representation.} Generalization experiments on the three-digit number recognition task show that a probe trained on addition generalizes very well to subtraction (peaking near 90\% accuracy) and strongly to multiplication (peaking near 80\%), and that its generalization to subtraction is even stronger than to multiplication. The relatively superior generalization to subtraction (i.e., marginally better performance at similarly high levels) may indicate that in this model’s representation learning, subtraction and addition share more direct or more readily transferable numerical representation elements. Although multiplication (conceptually a form of repeated addition, and likewise exhibiting good generalization here) also achieves a high degree of transfer, it falls slightly short of subtraction. This suggests that, despite substantial commonality, the computational mechanism the model learns for multiplication—or the representations it forms—retain some distinctive aspects that render its transfer a bit less effective than subtraction. Overall, this shows that the model’s learned numerical representations combine a certain level of abstraction and cross-operation universality with features that are specific to each arithmetic operation.

Thirdly, \textbf{the analogical nature of the model's "computation" process}. The sequential emergence and progressive refinement of features revealed in this study bear resemblance to the cognitive processes humans use to solve arithmetic problems, which involve breaking down steps and calculating progressively. From identifying the structure of the expression, to processing carries, calculating the sum, and finally confirming and expressing each digit, the model internally appears to be executing a structured, computation-like process, rather than simple pattern matching or memory retrieval.

\section{Conclusion}

Through a series of carefully designed probe experiments on the \textsc{LLaMA-3-8B-Instruct} model, and with comparative analyses on other representative models (detailed in Appendix \ref{sec:model_experiments}), this study systematically investigated their internal information processing mechanisms when performing addition operations. The main conclusions regarding \textsc{LLaMA-3-8B-Instruct} are as follows:

\paragraph{Sequential Information Processing Trajectory:}
Our findings indicate that when \textsc{LLaMA-3-8B-Instruct} processes addition operations, the decodability of its internal arithmetic features presents a coherent, sequential evolution pattern, outlining a multi-step information processing trajectory. Information at different levels, such as the recognition of formula structure, the emergence of features related to core operations (including carry and summation), and the numerical abstraction and output organization of the result, corresponds to representations that become clearly discernible at different, sequential depths of the network.

\paragraph{Formation of Explicit Representations:} At various nodes of this processing trajectory (corresponding to different network depths), the model forms internal representations of specific arithmetic features (such as carries, sum values, individual digits) that are clear, stable, and easily decodable by linear probes, and the strength of these representations changes dynamically as processing proceeds.

\paragraph{Support for "Computation-like" Process over "Rote Memorization":} The observed sequential emergence of features, the intrinsic connection between the operation results and their numerical attributes, the dynamic evolution of representations, and a certain degree of cross-task generalization capability collectively constitute strong evidence. This suggests that when \textsc{LLaMA-3-8B-Instruct} encounters simple addition problems within its training distribution, it is more inclined to perform a structured, "computation-like" process rather than primarily relying on large-scale pattern memorization.

Furthermore, comparative analyses with Qwen2.5-Math-7B-Instruct (see \ref{sec:model_experiments} for full details) revealed that the observed sequential processing trajectory and computation-like mechanisms are largely consistent across different model architectures, though with variations in the specific layers at which features emerge / certain model-specific strategies in representing numerical attributes
which helps to contextualize the findings from \textsc{LLaMA-3-8B-Instruct} and underscores the broader relevance of these information processing principles in LLMs.

In summary, the findings of this study deepen the understanding of the arithmetic capabilities of LLMs and provide empirical evidence for their complex internal information processing mechanisms, particularly revealing an ordered, computation-like internal information representation evolution path.

While some math-optimized models achieve higher raw accuracy on our addition benchmark, the four-stage trajectory we document is robust across architectures (Appendix~\ref{sec:model_experiments}) and is most cleanly expressed in \textsc{LLaMA-3-8B-Instruct}, which motivates our focus on this model for mechanistic analysis.

\section*{Limitations}
Our study offers a first layer-wise picture of how \textsc{LLaMA-3-8B-Instruct} handles multi-digit addition, but several caveats temper the strength of our claims.

\paragraph{Correlation \textit{vs.} causation.}
Our analysis uses \emph{correlational} tools—linear probes and the Logit Lens. These reveal that certain attributes are \emph{linearly decodable} from hidden states, but do not prove that the forward pass \emph{uses} those attributes, nor that the model implements a specific carry-propagation algorithm. Decodability can arise from distributed or incidental codes, and linear probes can overfit if not carefully regularized. Establishing \emph{dependence} on intermediate computations requires causal interventions (e.g., activation patching, targeted ablations, or circuit-level surgery), which we leave to future work.

\paragraph{Task scope.}
We study in-distribution, base-10 \emph{multi-digit addition} under short prompts of the form ``\texttt{Calculate: num1+num2 = }'' (with a trailing space) and evaluate 1–6 digits using exact-match metrics. This choice reflects data realism—decimal numerals dominate pretraining corpora—while avoiding confounds from distribution shift. We do not evaluate borrowing subtraction, long multiplication, mixed-base arithmetic, chain-of-thought prompting, or adversarial perturbations, and we therefore scope claims to base-10 addition under short prompts.

\paragraph{Model specificity.}
Our main analyses focus on \textsc{LLaMA-3-8B-Instruct}; we additionally report probe results for several 7B-scale models, including math-optimized variants, in Appendix~\ref{sec:model_experiments}. While the \emph{four-stage ordering} (structure $\rightarrow$ carries/sums $\rightarrow$ digit abstractions $\rightarrow$ output alignment) recurs qualitatively, the \emph{exact layer indices} and effect sizes vary across architectures and checkpoints. Thus, our claims concern \emph{ordering} rather than universal layer numbers. Broader generalization may be affected by architecture, scale, tokenizer, instruction tuning, and math-oriented fine-tuning.

\section*{Ethical Considerations}
This research focuses on investigating the internal information processing pathways of Large Language Models (LLMs) when performing basic multi-digit addition. The core of the study involves analyzing internal model activations using techniques like linear probing and logit lens to understand their computational mechanisms, rather than developing new applications or models.

We assess that this specific research work itself does not present significant ethical risks, for the following reasons:
\begin{enumerate}
    \item \textbf{Benign Nature of the Task:} The problem under investigation (multi-digit addition) is a fundamental and well-defined arithmetic task that does not involve sensitive content, personal data, or socially charged issues that might introduce bias.
    \item \textbf{Non-sensitive Data:} The datasets used in this study (e.g., prompts of the form "Calculate: num1+num2 = " and corresponding arithmetic problems) are synthetically constructed for the research purpose and do not contain any personally identifiable information or other sensitive information.
    \item \textbf{Transparency in Research Aim:} This study aims to enhance the understanding of the internal workings of LLMs, contributing to model interpretability and transparency. 
\end{enumerate}
While Large Language Models as a broader technology may raise various ethical considerations (such as bias, misuse, environmental impact, etc.), the scope of this research is limited to a mechanistic investigation of arithmetic capabilities in specific models (e.g., LLaMA-3-8B-Instruct). It does not directly engage with these broader ethical issues. Therefore, we believe the direct ethical risks associated with the current work are minimal and manageable.

\section*{Acknowledgements}
This work was supported in part by the Chongqing Key Lab of Cognitive Intelligence and  Intelligent Finance
and the College of Computer and Information Science at Chongqing Normal University.
Computational resources were provided by the Chongqing Key Lab of Cognitive Intelligence and Intelligent Finance.
This research was funded by the Chongqing Graduate Student Research Innovation Project (Grant No.\ YKC24001).


\bibliography{acl_latex.bib}

\appendix

\label{sec:appendix}

\section{Detailed Model Performance Metrics}\label{sec:Metrics} 

This appendix presents a detailed summary of the performance metrics for the different language models evaluated in this study. Table~\ref{tab:model_performance_appendix} offers a comparative analysis of these models based on their accuracy in specific numerical prediction tasks.

Each row in the table corresponds to a distinct language model, identified under the "Model ID" column. The "Overall Accuracy" column provides a general measure of each model's performance across all evaluated tasks. The subsequent columns, labeled from "1-digit" through "6-digit," specify the model's accuracy in correctly predicting numerical values composed of one to six digits, respectively. All accuracy figures presented in the table are expressed as percentages. This data provides insight into the capabilities and limitations of each model concerning tasks that require numerical precision and understanding of magnitude.

\begin{table*}[htbp!]
  \centering
  \begin{tabular}{@{}l*{7}{c}@{}}
    \hline
    \textbf{Model ID} & \textbf{Overall Accuracy} & \textbf{1-digit} & \textbf{2-digit} & \textbf{3-digit} & \textbf{4-digit} & \textbf{5-digit} & \textbf{6-digit} \\
    \hline
    \textsc{LLaMA-3-8B-Instruct} & 97.29\% & 100.00\% & 100.00\% & 99.70\% & 95.51\% & 93.21\% & 95.31\% \\
    Mistral-7B-Instruct-v0.2 & 84.99\% & 98.81\% & 83.05\% & 89.03\% & 84.63\% & 79.94\% & 74.45\% \\
    Qwen3-8B & 63.98\% & 92.18\% & 79.52\% & 52.09\% & 57.57\% & 54.00\% & 48.53\% \\
    Qwen2.5-7B-Instruct & 60.61\% & 77.31\% & 71.88\% & 67.90\% & 37.03\% & 50.50\% & 58.98\% \\

    \hline
  \end{tabular}
  \caption{\label{tab:model_performance_appendix}
Overall Model Performance Summary. This table shows the accuracy of different models across various digit prediction tasks. The target for each group was to collect 4000 samples, though some groups had fewer (for example, the 1-digit group only contained 55 samples).
  }
\end{table*}

\section{General Experimental Setup}\label{sec:AppGeneral}

\subsection{Language Model}\label{sec:app-language-model}
All experiments used the \texttt{\textsc{LLaMA-3-8B-Instruct}} model.  Model parameters
were kept frozen for all probing runs.  The network has 32 transformer layers.

\subsection{Input Representation for Probes}\label{sec:app-input-repr}
For each of the 32 layers an independent diagnostic probe was trained.
The probe input was the hidden state of the \emph{last} token of the prompt
\verb|Calculate: num1 + num2 =|—namely the trailing space after the ``='' sign.

\subsection{Probe Architecture}\label{sec:app-probe-arch}
Unless otherwise stated (see \autoref{sec:app-exp-details}), every probe
consisted of
\begin{itemize}
  \item a single linear classification layer, followed by
  \item a \textsc{Softmax} activation.
\end{itemize}
This yields a probability distribution over the task-specific classes.

\subsection{Training and Evaluation}\label{sec:app-train-eval}
\begin{itemize}
  \item \textbf{Data splitting}: Each dataset was split 80 \%/20 \% into
        train/test using stratified sampling to preserve class ratios.
  \item \textbf{Replications}: All experiments were run five times
        with random seeds 42-46; reported scores are the mean over runs.
\end{itemize}

\section{Experiment-Specific Details}\label{sec:app-exp-details}

\subsection{Formula Structure Representation (Addition-Order Experiments)}
\label{sec:app-add-order}

\paragraph{Task.}
Distinguish three expression types for two-digit integers
\(a,b\in[10,99]\) with \(a>b\):
\begin{enumerate}
  \item \(a+b\) (e.g.\ \(45+23\))
  \item \(b+a\) (e.g.\ \(23+45\))
  \item \(a+a\) (e.g.\ \(66+66\))
\end{enumerate}

\paragraph{Training data.}
\begin{itemize}
  \item Type 1: 5000 random samples
  \item Type 2: 5000 random samples
  \item Type 3: 80\% of the 90 unique pairs
\end{itemize}

\paragraph{Generalization tests.}
Three extra test sets mirror the above but use
one-, three-, and four-digit addends, respectively.

\subsection{Emergence of Core Computational Features (Arithmetic-Results Experiments)}
\label{sec:app-sum-class}

\paragraph{Task.} Predict the exact sum; each class is one sum value.

\paragraph{Dataset generation.}
\begin{itemize}
  \item 400 examples per class with unconstrained addends.
  \item Five groups of sums: 500-509, 600-609, 700-709, 800-809, 900-909.
\end{itemize}

\subsection{Carry-Signal Experiment}\label{sec:app-carry}

\paragraph{Task.} Binary classification—detect a carry in the units,
tens, or hundreds place of three-digit additions.

\paragraph{Datasets.}
Three balanced sets of 1 000 problems each (one per decimal place).
A single problem may appear in several sets if its carries differ by place.

\subsection{Numerical Abstraction of Results}\label{sec:app-digit-id}

\paragraph{Digit-identification task.}
Given a sum that is three digits long, predict its
hundreds, tens, and units digits (10-way classification each).
The dataset contains 1 000 two-addend problems.

\paragraph{Hundreds-digit generalization.}
A probe trained on hundreds-digit identification was also tested on
subtraction and multiplication datasets whose results are mostly
three-digit numbers, generated with the same principles.

\subsection{Logit Lens Analysis}\label{sec:app_logit_lens}

\paragraph{Task.}
Find the earliest transformer layer whose output distribution ranks
the correct sum token at top-1 (logit lens).

\paragraph{Dataset.}
1000 randomly generated problems:
two three-digit addends yielding a three-digit sum.

\section{Model Experiment Results}\label{sec:model_experiments}

\subsection{Arithmetic Structure Recognition Experiments}\label{sec:arith-structure}
\begin{figure}[H]
\includegraphics[width=\columnwidth]{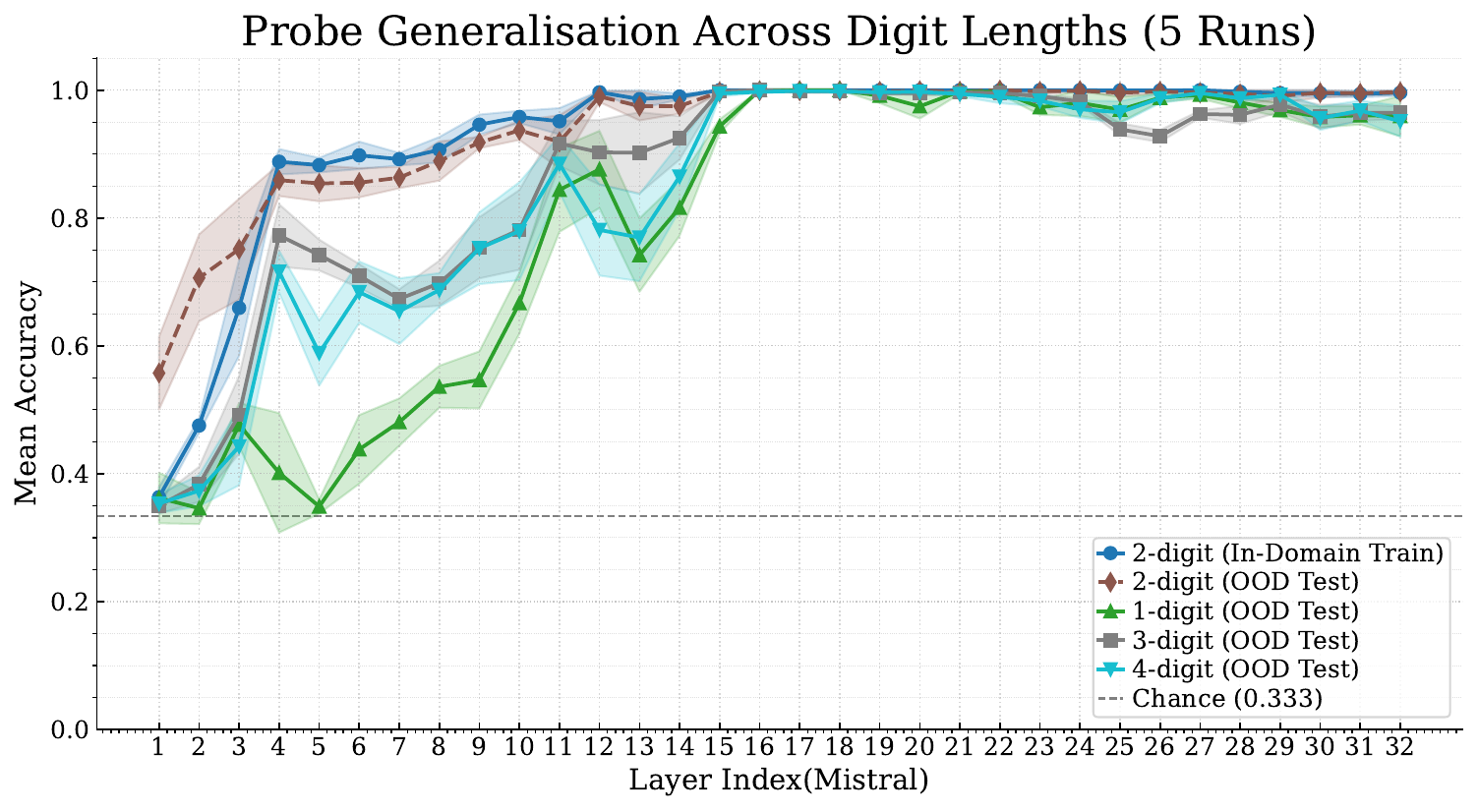}
\caption{Mistral-7B-Instruct: Probe trained on a two-digit addition dataset; accuracy by digit position.}
\label{fig:mistral_addition_order}
\end{figure}

\begin{figure}[H]
\includegraphics[width=\columnwidth]{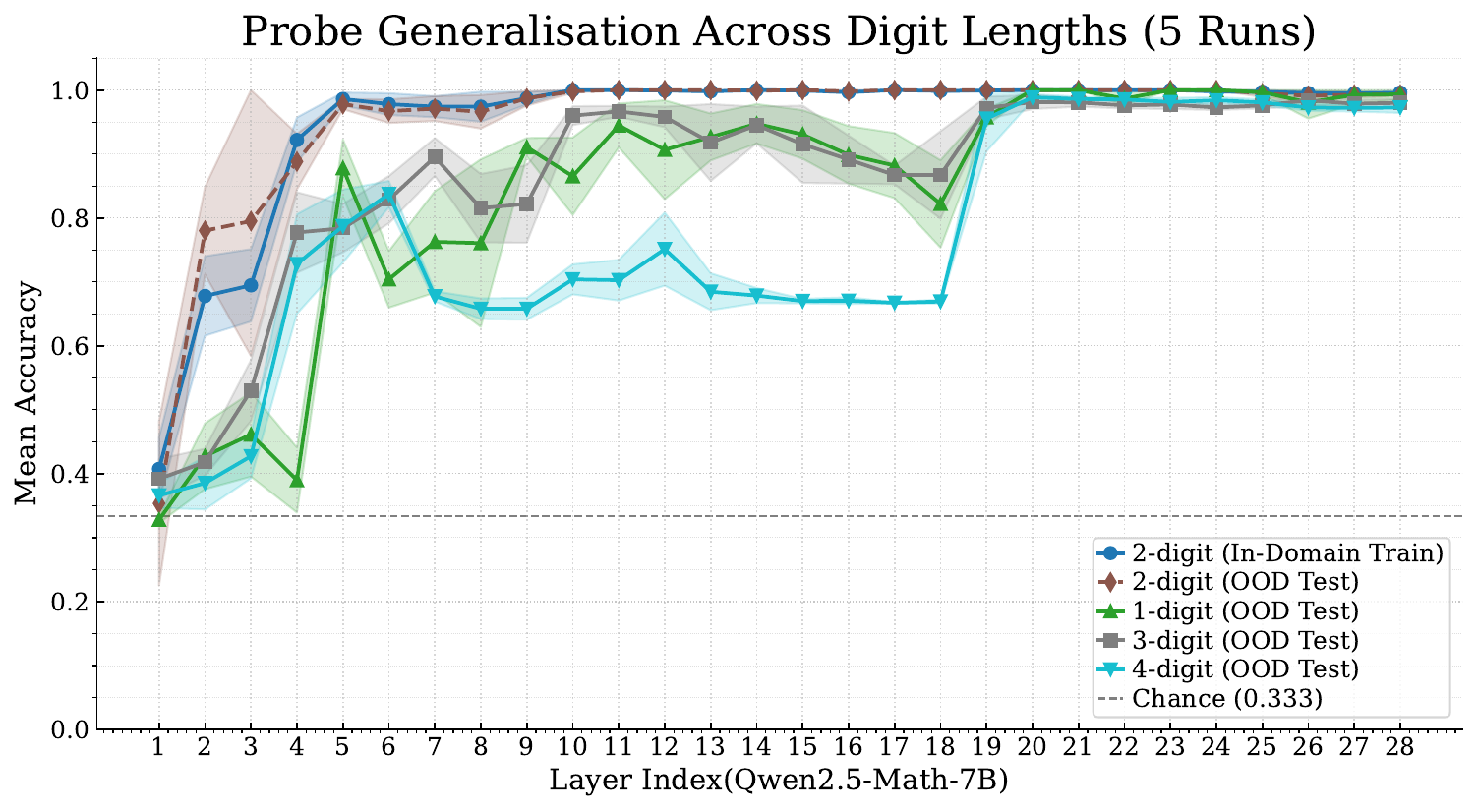}
\caption{Qwen2.5-Math-7B: Probe trained on a two-digit addition dataset; accuracy by digit position.}
\label{fig:qwen_addition_order}
\end{figure}

\begin{figure}[H]
\includegraphics[width=\columnwidth]{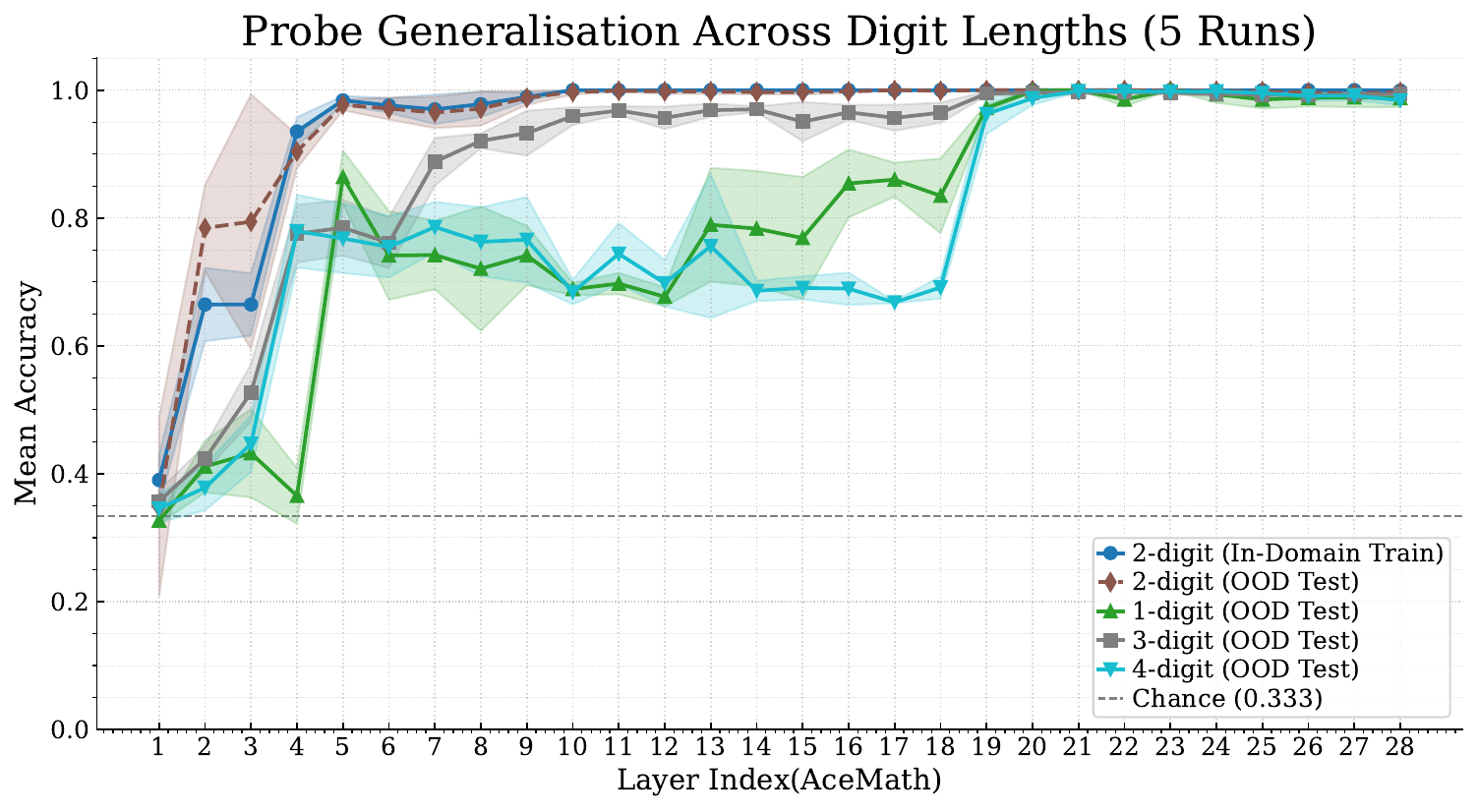}
\caption{AceMath: Probe trained on a two-digit addition dataset; accuracy by digit position.}
\label{fig:acemath_addition_order}
\end{figure}

\subsection{Arithmetic Results Experiments}\label{sec:arith-results}
\begin{figure}[H]
\includegraphics[width=\columnwidth]{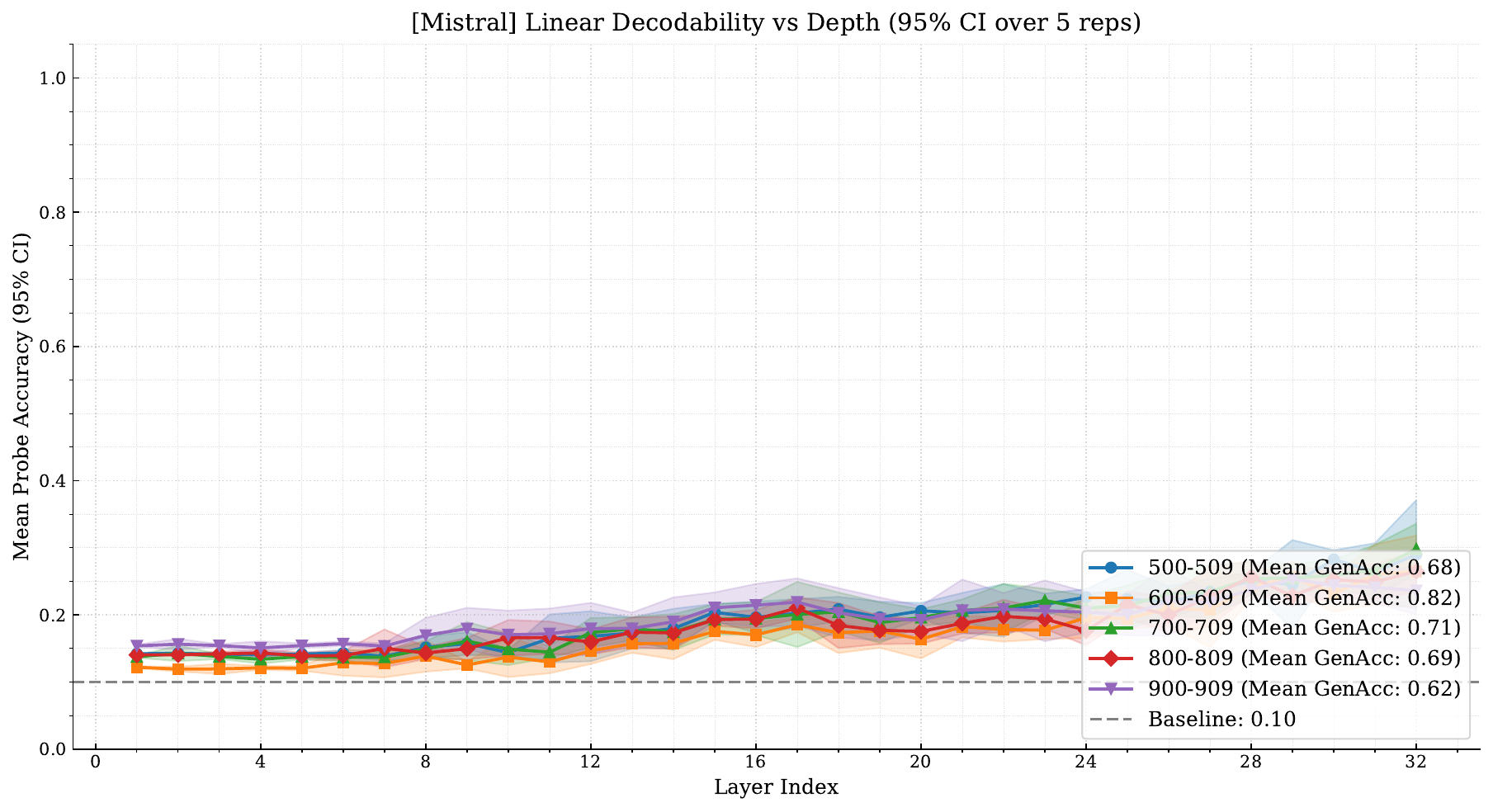}
\caption{Mistral-7B-Instruct: Layer-wise accuracy on arithmetic detection across numerical ranges (500--509 to 900--909). Note the anomaly: for the constructed dataset ``$a+b=500$'', correct samples are only $\sim$10\% of the total, so results for this slice are not displayed normally.}
\label{fig:mistral_arith_results}
\end{figure}

\begin{figure}[H]
\includegraphics[width=\columnwidth]{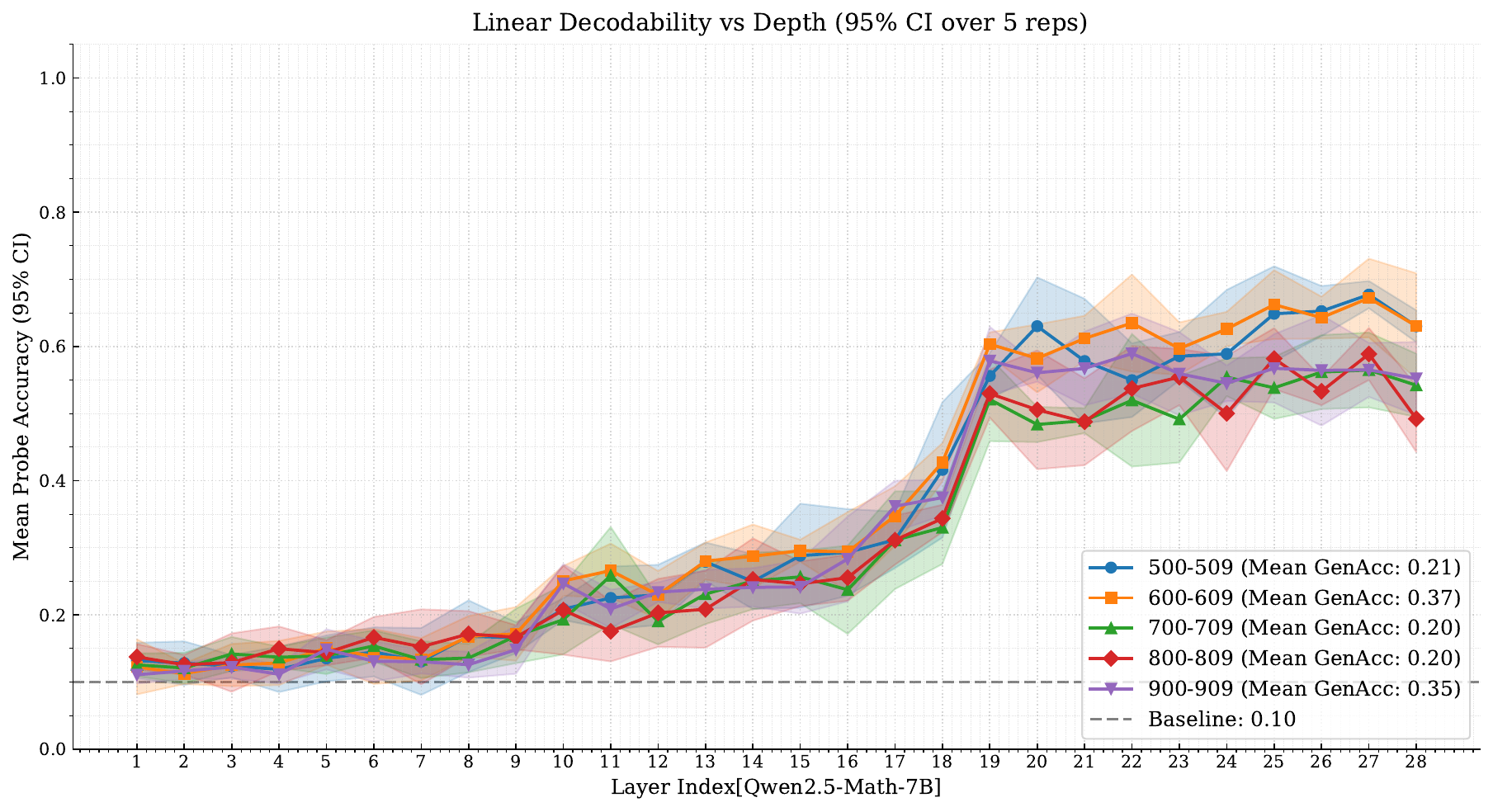}
\caption{Qwen2.5-Math-7B: Layer-wise accuracy on arithmetic detection across numerical ranges (500--509 to 900--909). The same anomaly applies for the ``$a+b=500$'' slice (correct $\sim$10\%).}
\label{fig:qwen_arith_results}
\end{figure}

\begin{figure}[H]
\includegraphics[width=\columnwidth]{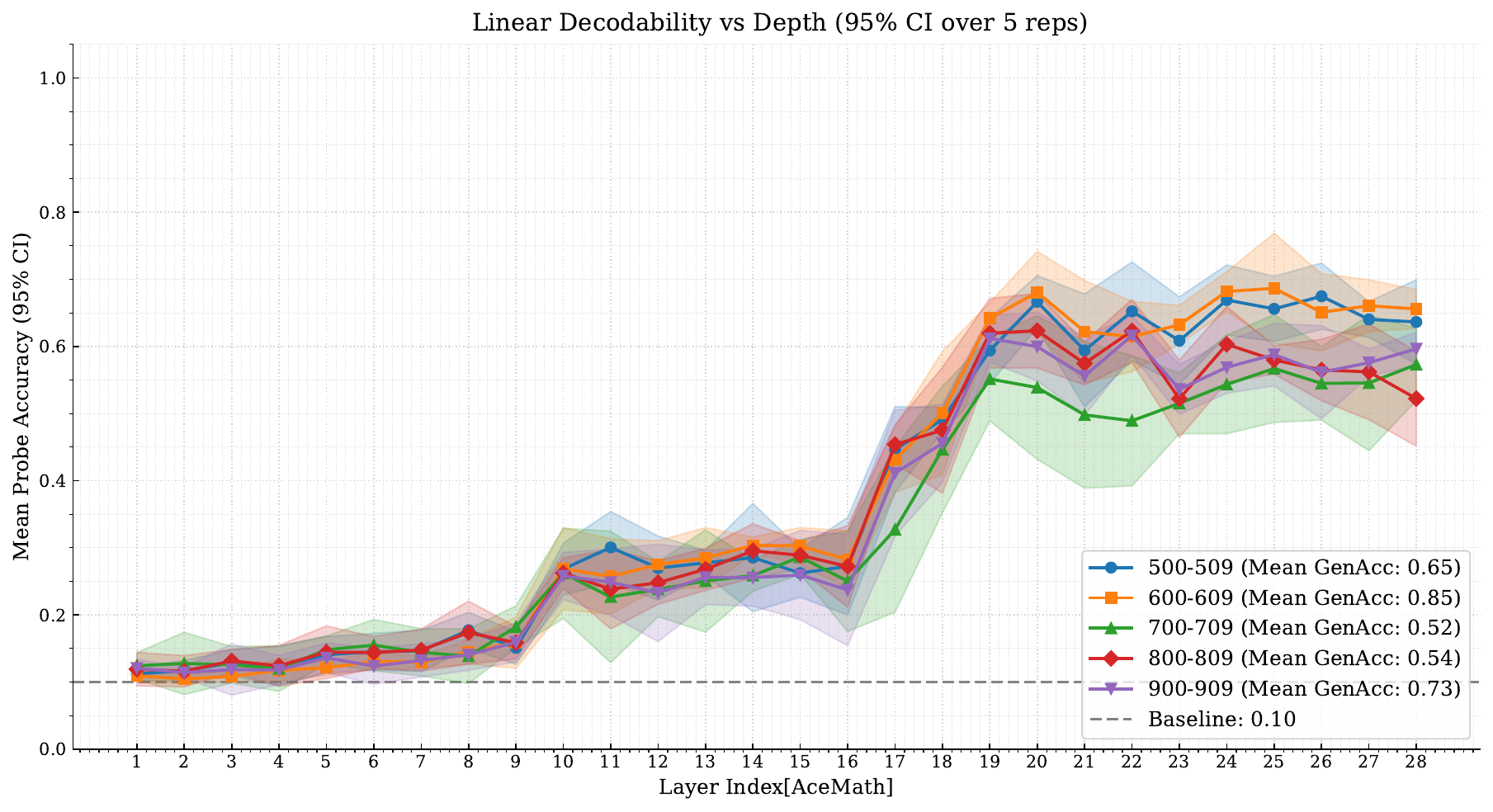}
\caption{AceMath: Layer-wise accuracy on arithmetic detection across numerical ranges (500--509 to 900--909). The same anomaly applies for the ``$a+b=500$'' slice (correct $\sim$10\%).}
\label{fig:acemath_arith_results}
\end{figure}

\subsection{Carry Signal Experiment}\label{sec:carry-signal}
\begin{figure}[H]
\includegraphics[width=\columnwidth]{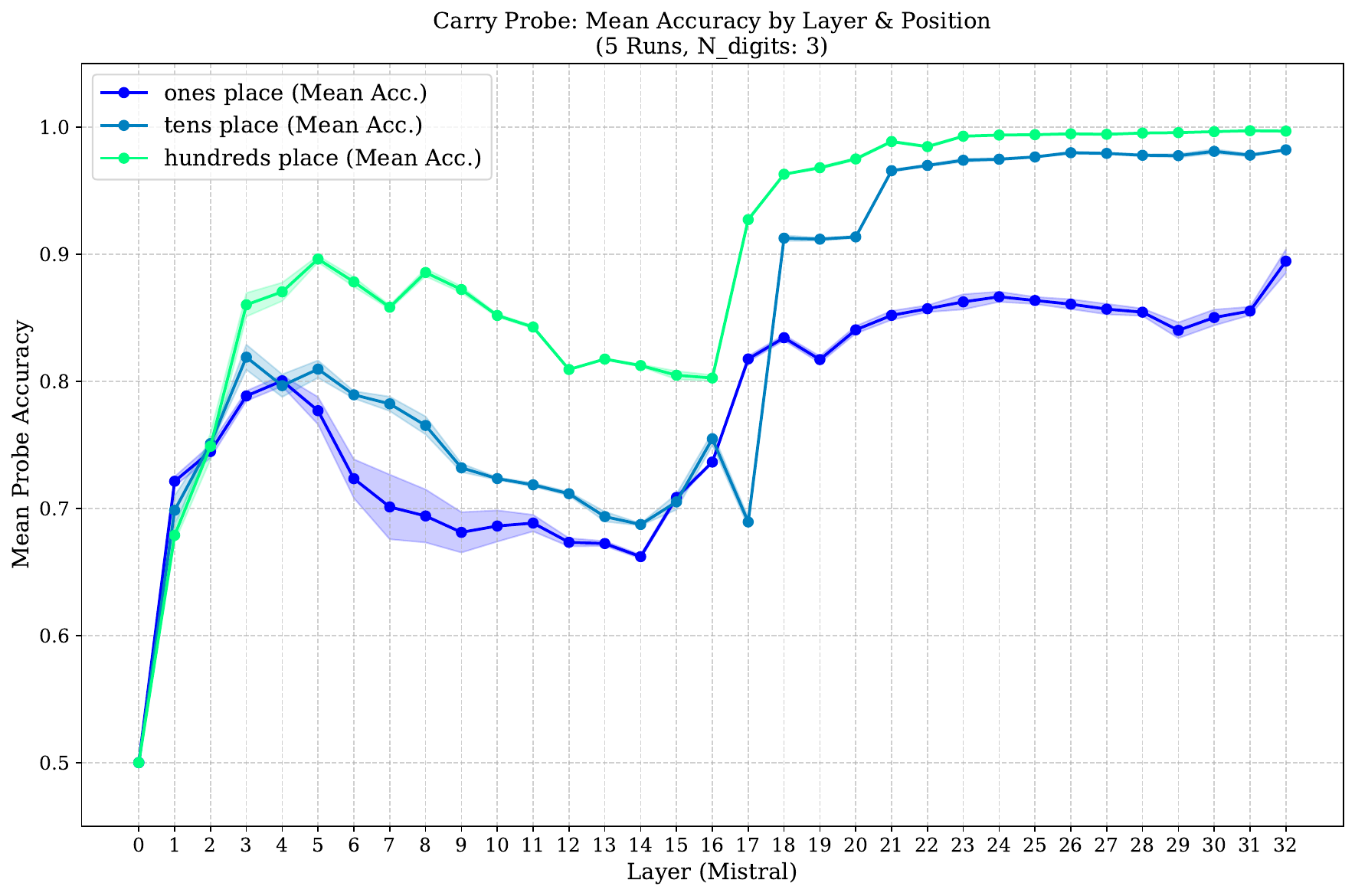}
\caption{Mistral-7B-Instruct: Carry detection---recognition accuracy of ones, tens, and hundreds at each layer.}
\label{fig:mistral_carry}
\end{figure}

\begin{figure}[H]
\includegraphics[width=\columnwidth]{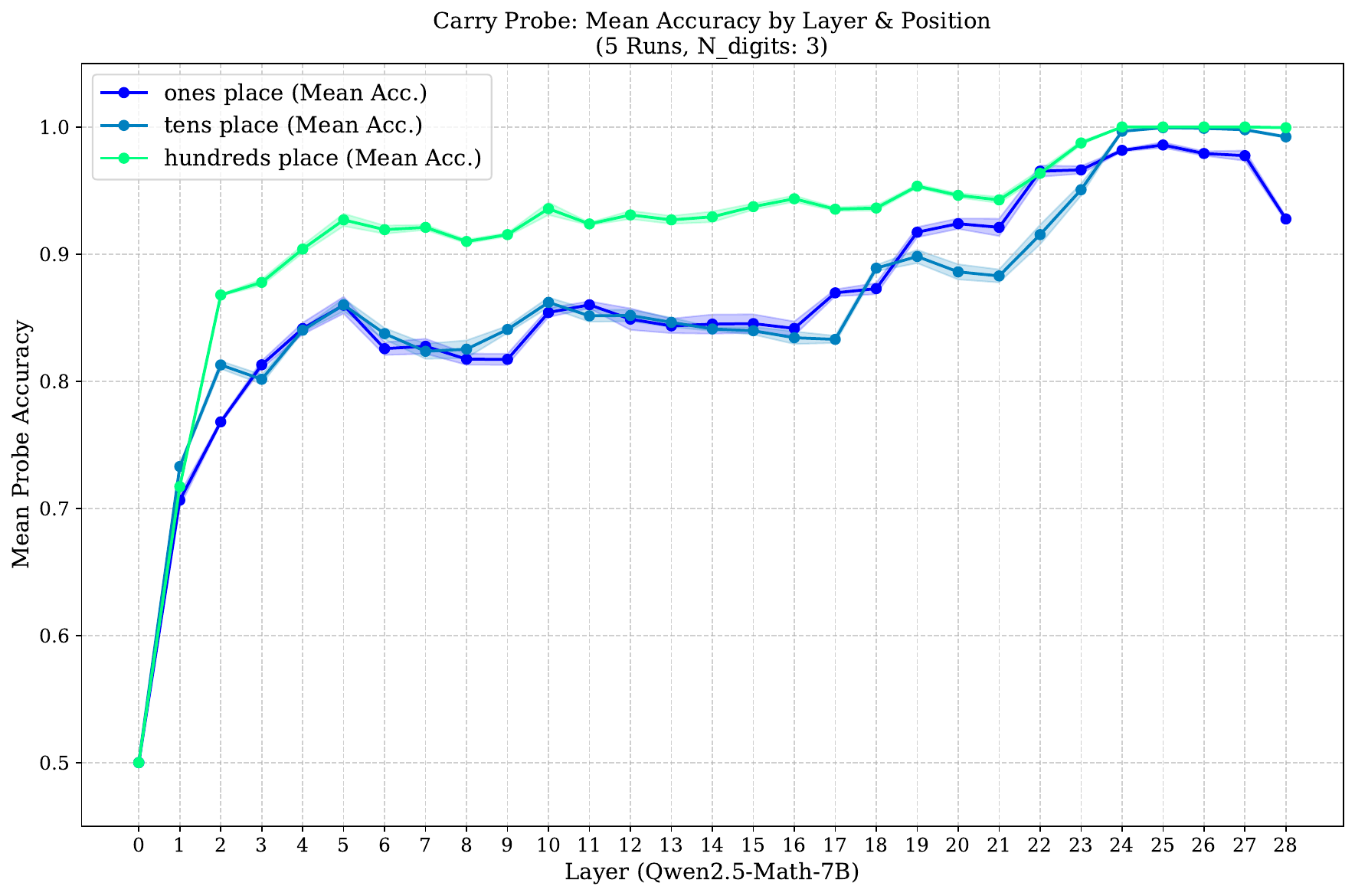}
\caption{Qwen2.5-Math-7B: Carry detection---recognition accuracy of ones, tens, and hundreds at each layer.}
\label{fig:qwen_carry}
\end{figure}

\begin{figure}[H]
\includegraphics[width=\columnwidth]{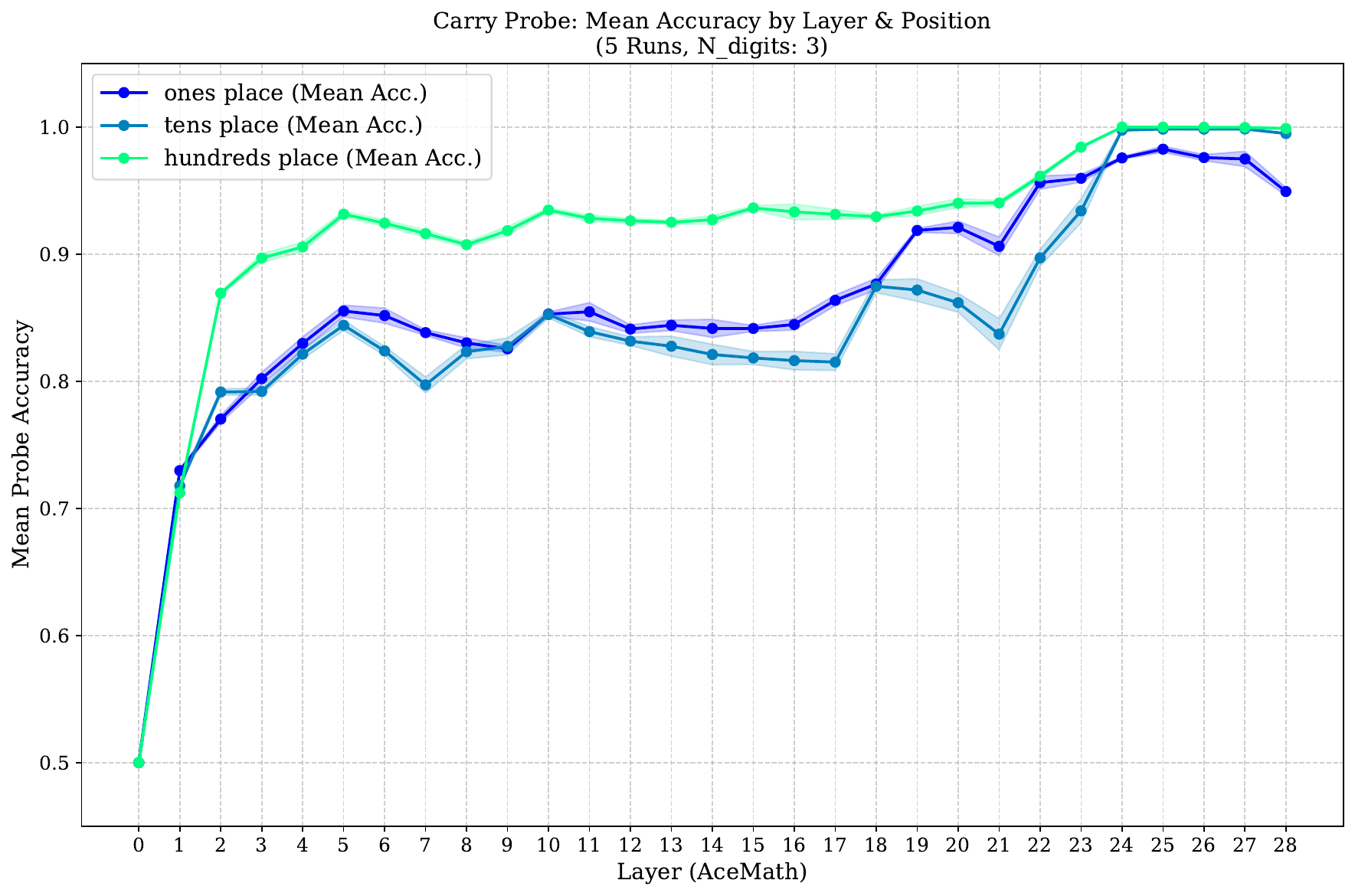}
\caption{AceMath: Carry detection---recognition accuracy of ones, tens, and hundreds at each layer.}
\label{fig:acemath_carry}
\end{figure}

\subsection{Determining Digits of Addition Results}\label{sec:digit-det}
\begin{figure}[H]
\includegraphics[width=\columnwidth]{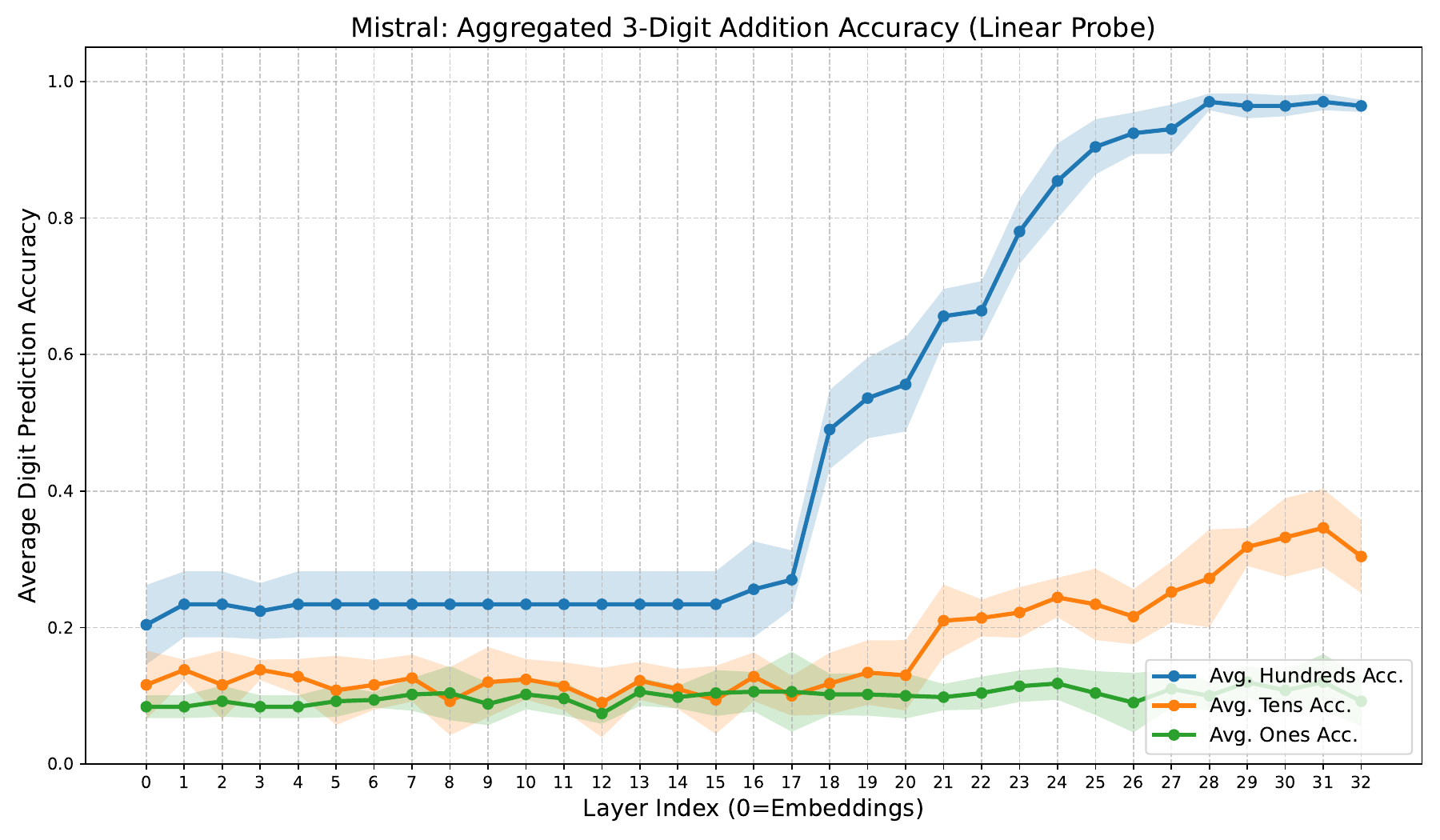}
\caption{Mistral-7B-Instruct: Layer-wise recognition accuracy for ones, tens, and hundreds of the sum.}
\label{fig:mistral_digit_det}
\end{figure}

\begin{figure}[H]
\includegraphics[width=\columnwidth]{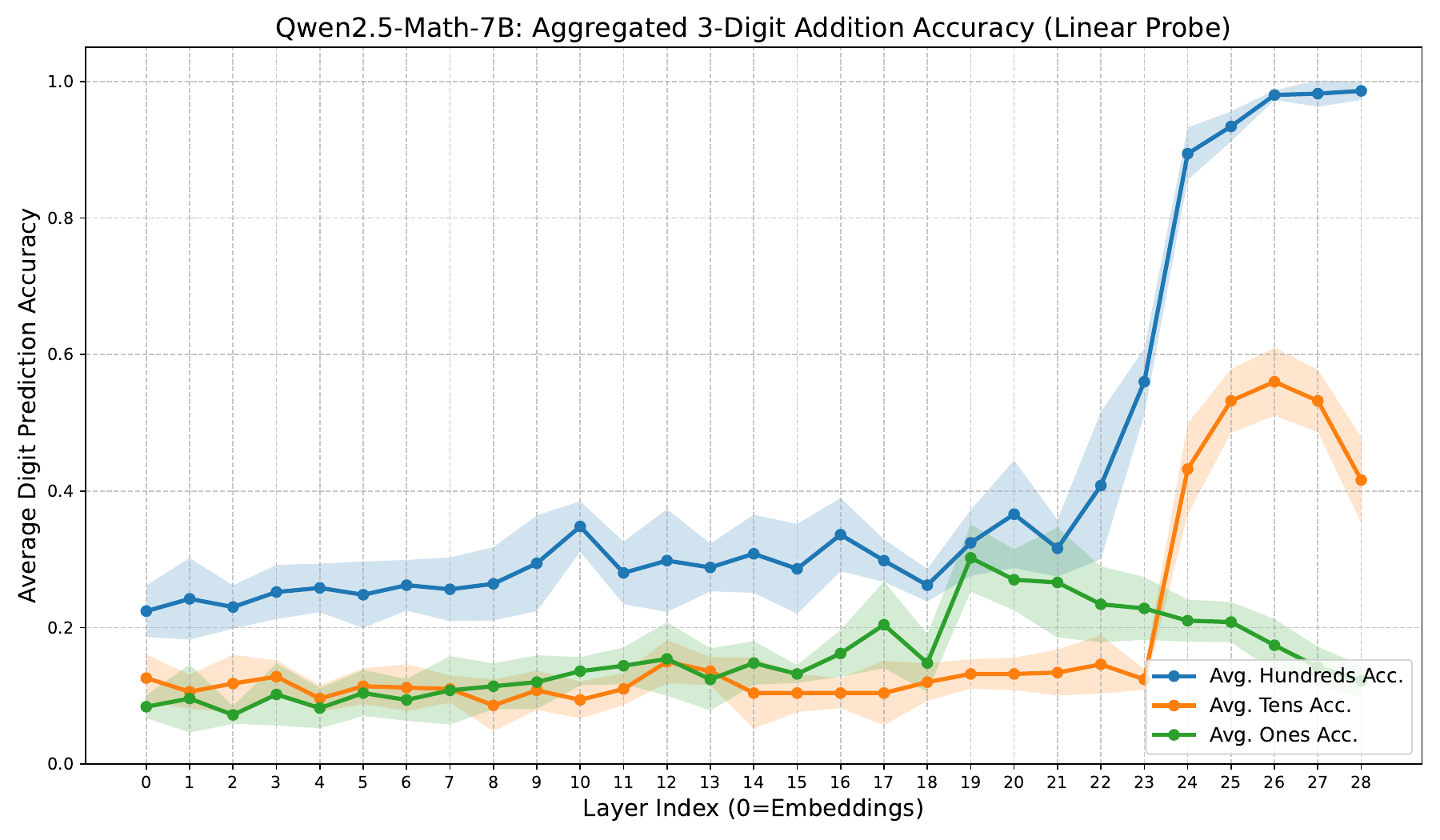}
\caption{Qwen2.5-Math-7B: Layer-wise recognition accuracy for ones, tens, and hundreds of the sum.}
\label{fig:qwen_digit_det}
\end{figure}

\begin{figure}[H]
\includegraphics[width=\columnwidth]{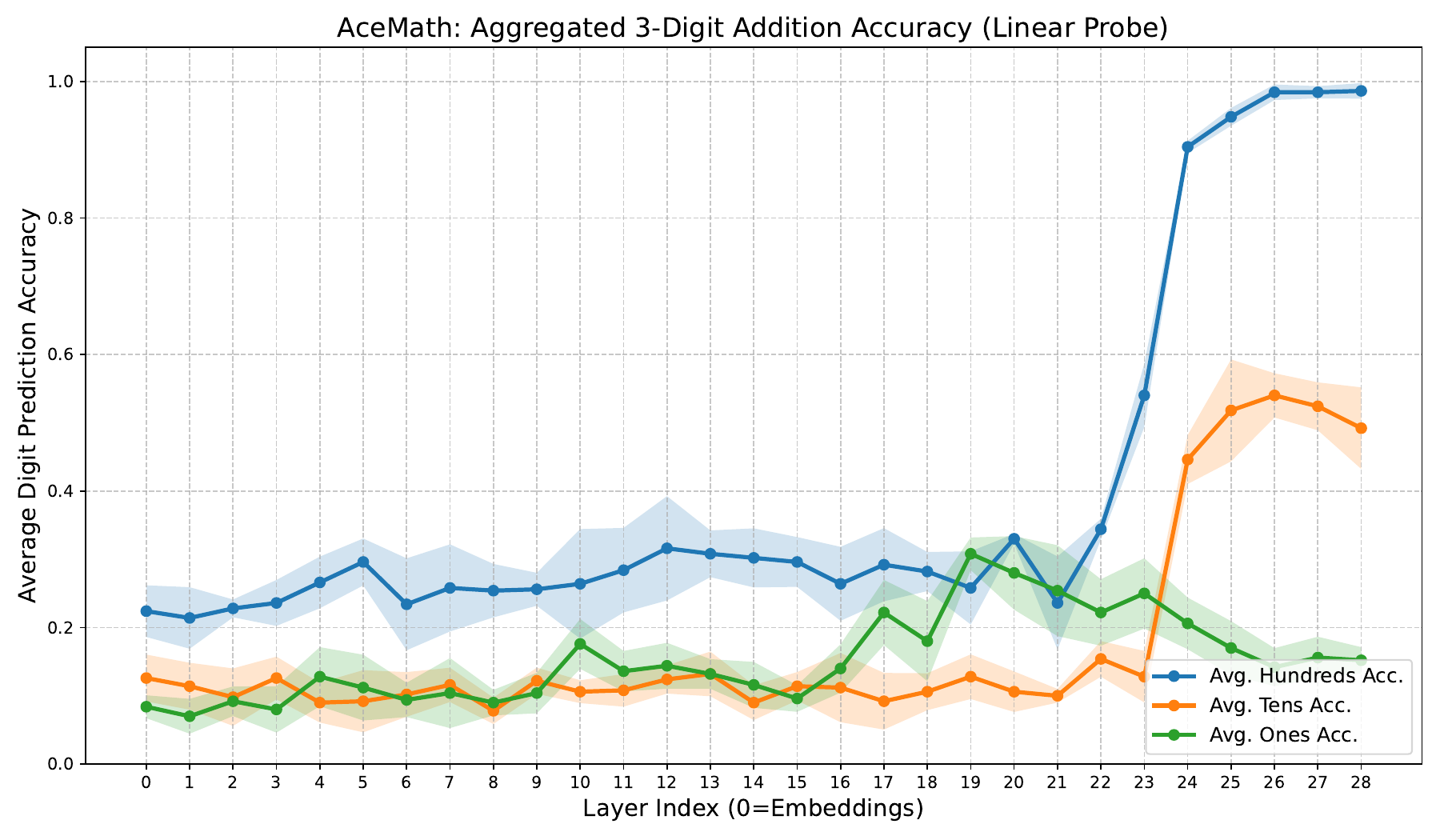}
\caption{AceMath: Layer-wise recognition accuracy for ones, tens, and hundreds of the sum.}
\label{fig:acemath_digit_det}
\end{figure}

\subsection{Numerical Abstraction Experiment}\label{sec:numerical-abstraction}
\begin{figure}[H]
\includegraphics[width=\columnwidth]{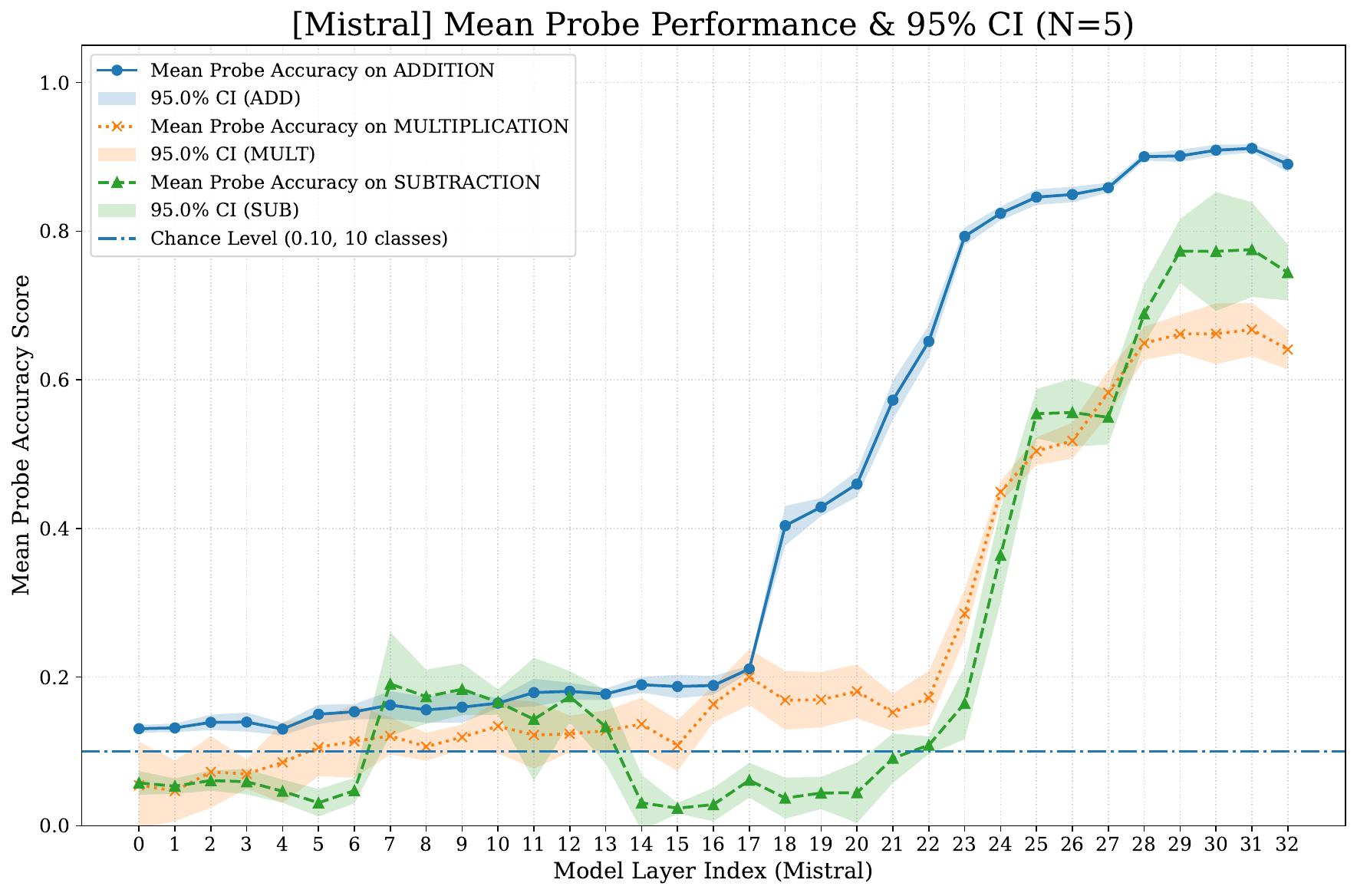}
\caption{Mistral-7B-Instruct: Hundreds-digit detection accuracy when applying an addition-trained probe to addition, multiplication, and subtraction test sets.}
\label{fig:mistral_hundreds_generalization}
\end{figure}

\begin{figure}[H]
\includegraphics[width=\columnwidth]{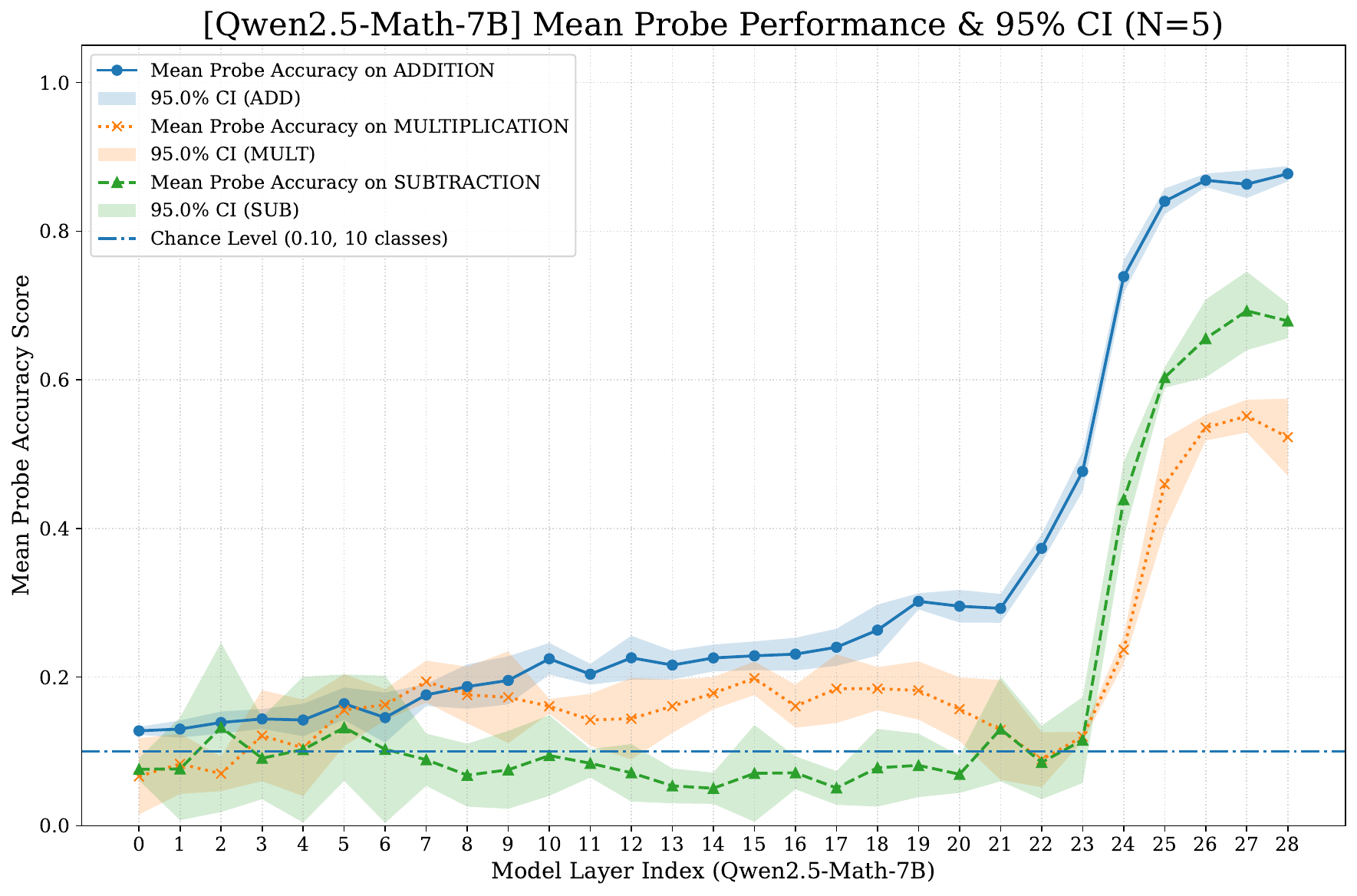}
\caption{Qwen2.5-Math-7B: Hundreds-digit detection accuracy when applying an addition-trained probe to addition, multiplication, and subtraction test sets.}
\label{fig:qwen_hundreds_generalization}
\end{figure}

\begin{figure}[H]
\includegraphics[width=\columnwidth]{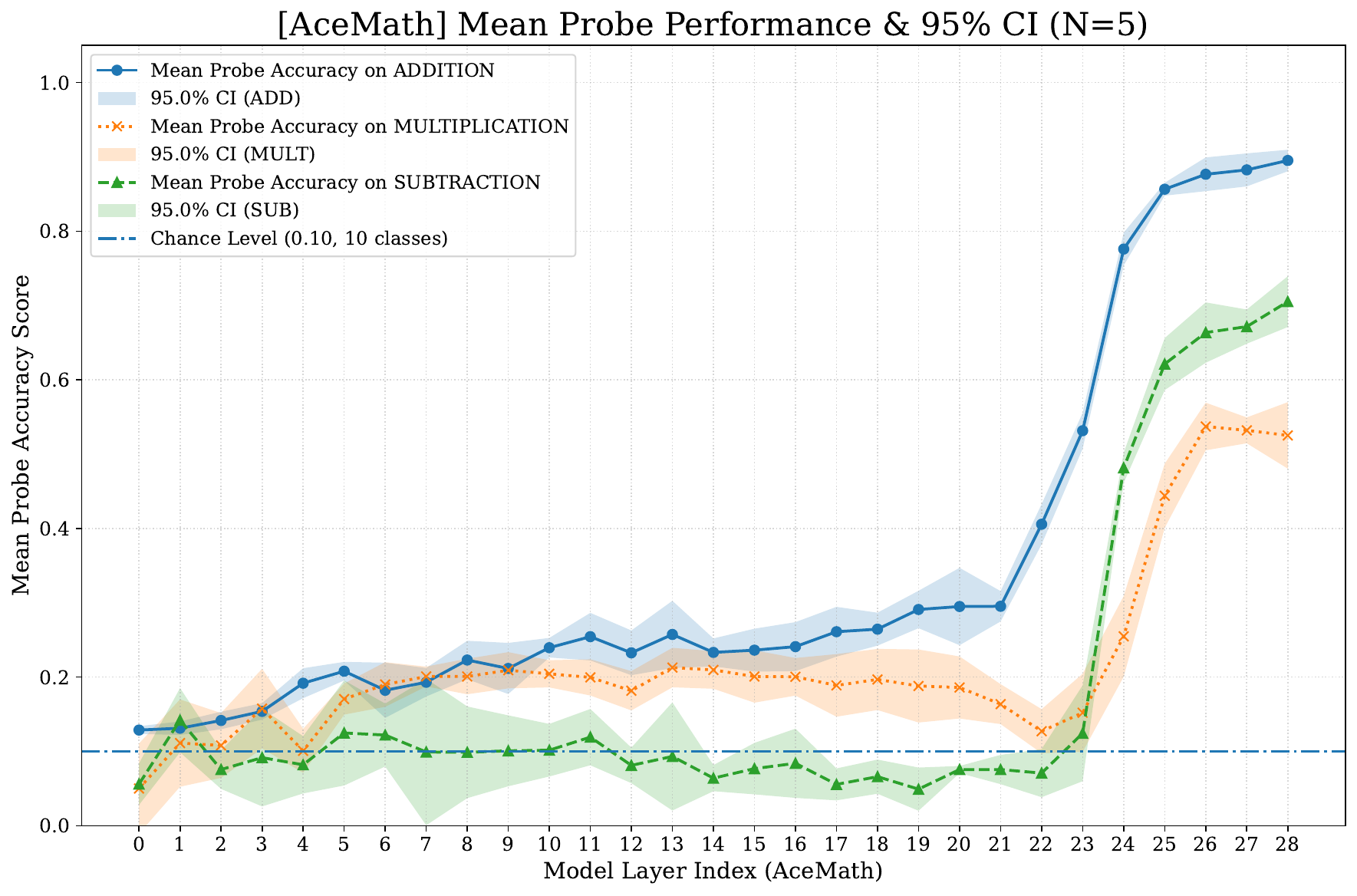}
\caption{AceMath: Hundreds-digit detection accuracy when applying an addition-trained probe to addition, multiplication, and subtraction test sets.}
\label{fig:acemath_hundreds_generalization}
\end{figure}

\subsection{First-Correct-Token Emergence}\label{sec:first-correct-token}
\begin{figure}[H]
\includegraphics[width=\columnwidth]{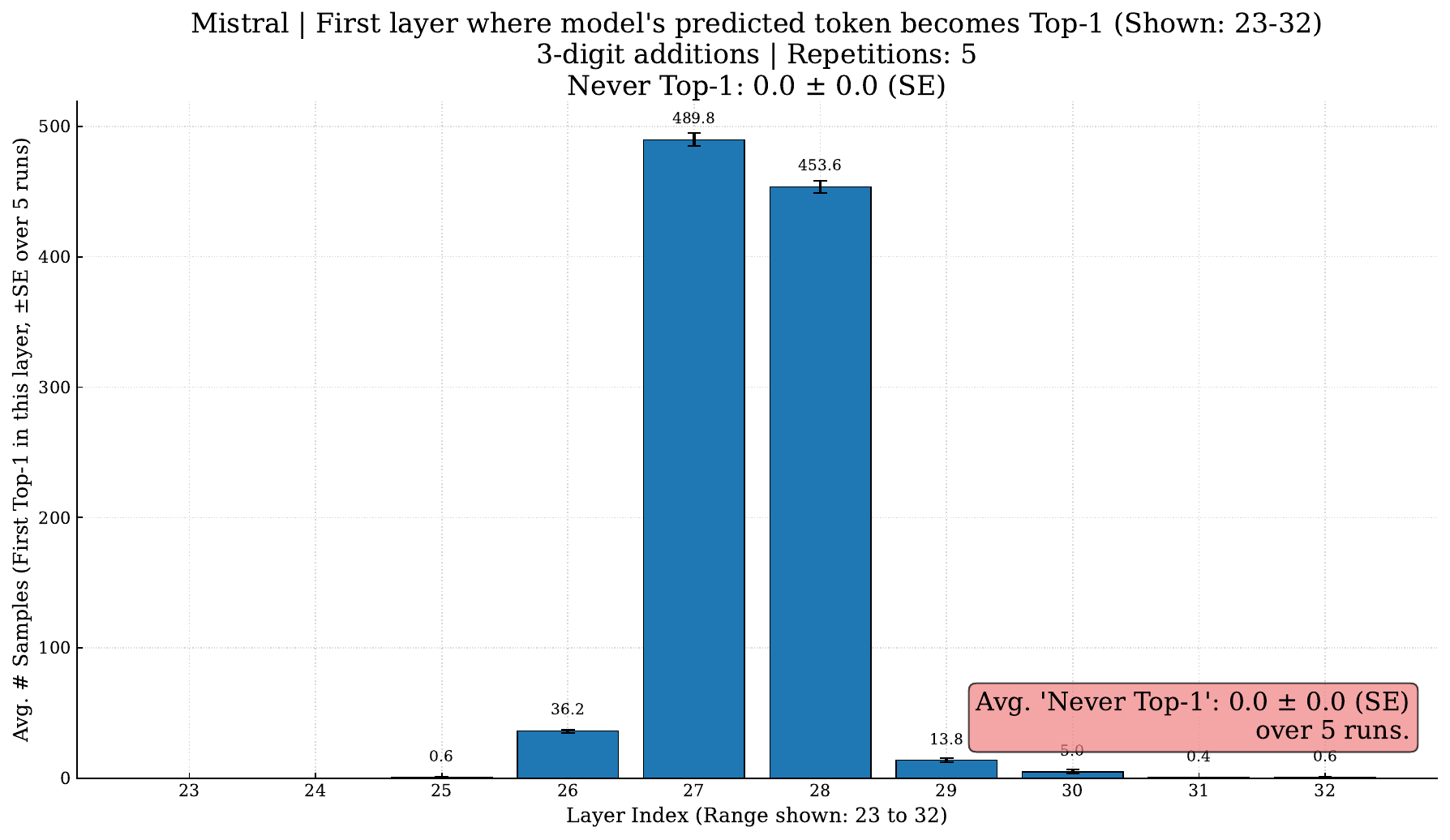}
\caption{Mistral-7B-Instruct: Logit Lens on layers 23--32; distribution over $N{=}1000$ samples where the correct sum token first becomes top-1.}
\label{fig:mistral_logitlens}
\end{figure}

\begin{figure}[H]
\includegraphics[width=\columnwidth]{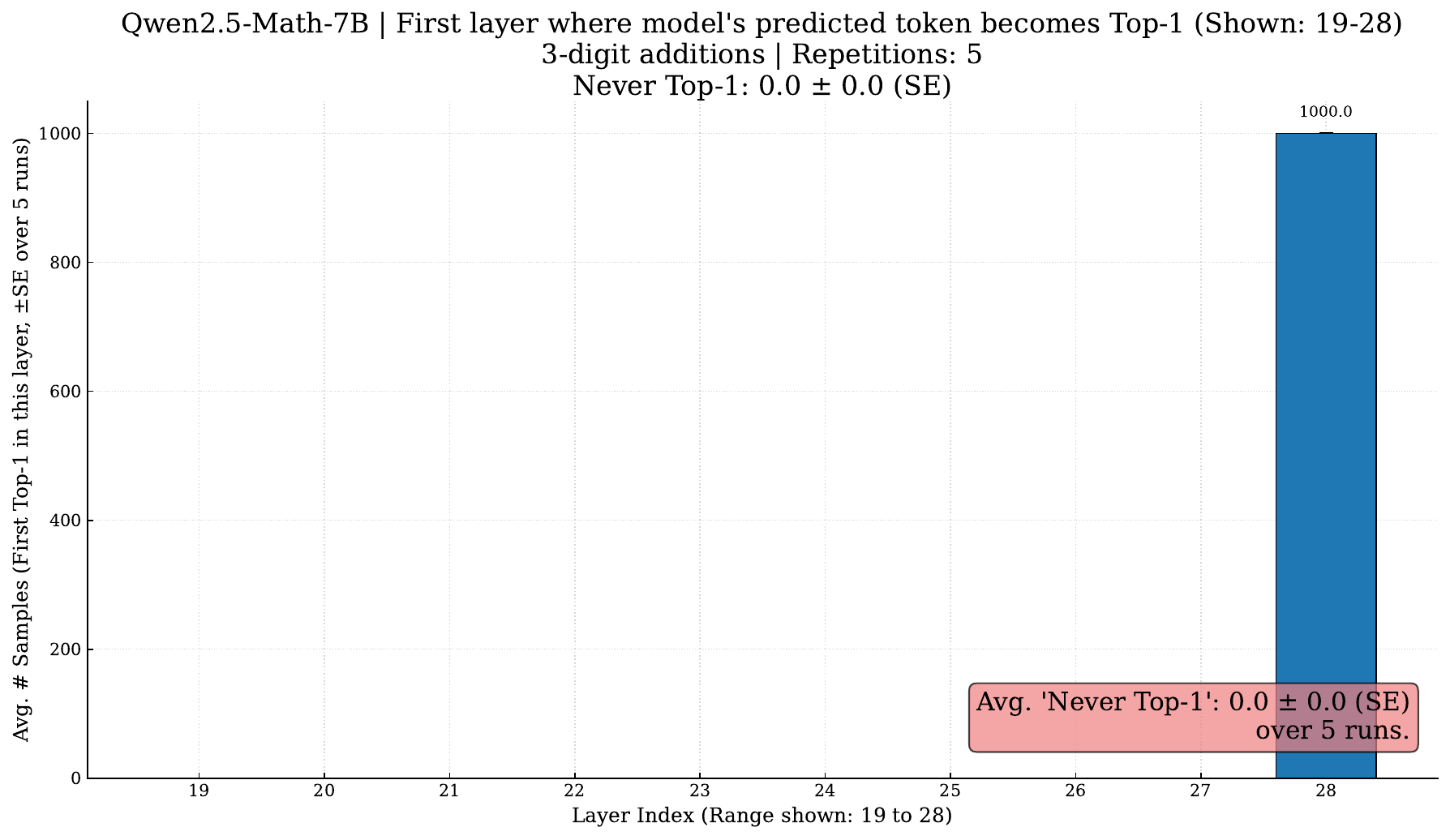}
\caption{Qwen2.5-Math-7B: Logit Lens analysis (layers 23--32), distribution of first top-1 correct token.}
\label{fig:qwen_logitlens}
\end{figure}

\begin{figure}[H]
\includegraphics[width=\columnwidth]{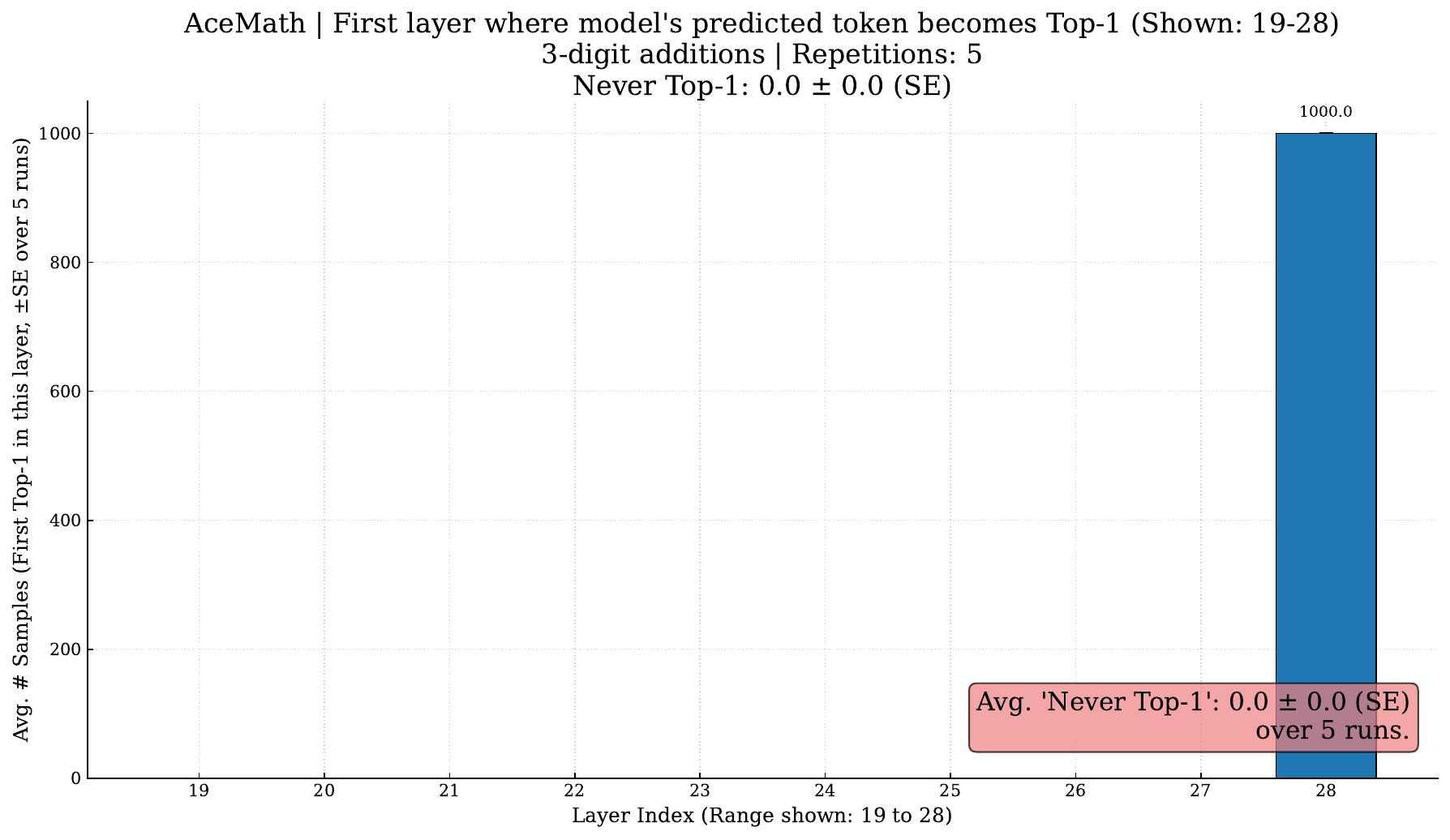}
\caption{AceMath: Logit Lens analysis, distribution of first top-1 correct token across the last 10 layers.}
\label{fig:acemath_logitlens}
\end{figure}

\end{document}